\documentclass{article}
\usepackage{times}
\usepackage[accepted]{icml2020}

\usepackage{booktabs}
\usepackage{graphicx}
\usepackage{epigraph}
\usepackage{subcaption}
\usepackage{amsmath}
\usepackage{amsfonts}
\usepackage{amssymb}
\usepackage{algorithm}
\usepackage{algorithmic}
\usepackage{afterpage}
\usepackage{ragged2e}

\usepackage{amsmath,amsfonts,bm}

\def\eqref#1{equation~\ref{#1}}
\def\1{\bm{1}}

\DeclareMathAlphabet{\mathsfit}{\encodingdefault}{\sfdefault}{m}{sl}
\SetMathAlphabet{\mathsfit}{bold}{\encodingdefault}{\sfdefault}{bx}{n}

\def\gA{{\mathcal{A}}}

\def\gS{{\mathcal{S}}}

\newcommand{\E}{\mathbb{E}}

\usepackage{hyperref}
\usepackage{url}
\usepackage{verbatim}
\usepackage{multicol}

\usepackage{natbib}

\DeclareUnicodeCharacter{2212}{-}

\icmltitlerunning{Striving for simplicity and performance in off-policy DRL}
\begin{document}

\twocolumn[
\icmltitle{Striving for Simplicity and Performance in Off-Policy DRL:\\ Output Normalization and Non-Uniform Sampling}

\icmlsetsymbol{equal}{*}

\begin{icmlauthorlist}
\icmlauthor{Che Wang}{equal,nyu,nyush}
\icmlauthor{Yanqiu Wu}{equal,nyu,nyush}
\icmlauthor{Quan Vuong}{ucsd}
\icmlauthor{Keith Ross}{nyu,nyush}
\end{icmlauthorlist}

\icmlaffiliation{nyu}{Department of Computer Science, New York University, New York, NY, USA}
\icmlaffiliation{nyush}{Department of Computer Science, NYU Shanghai, Shanghai, China}
\icmlaffiliation{ucsd}{Department of Computer Science, University of California San Diego, San Diego, CA, USA}

\icmlcorrespondingauthor{Keith Ross}{keithwross@nyu.edu}

\icmlkeywords{Deep Reinforcement Learning, Reinforcement Learning, ICML}

\vskip 0.3in]

\printAffiliationsAndNotice{\icmlEqualContribution}

\begin{abstract}
We aim to develop off-policy DRL algorithms that not only exceed state-of-the-art performance but are also simple and minimalistic. For standard continuous control benchmarks, Soft Actor-Critic (SAC), which employs entropy maximization, currently provides state-of-the-art performance. We first demonstrate that the entropy term in SAC addresses action saturation due to the bounded nature of the action spaces, with this insight, we propose a streamlined algorithm with a simple normalization scheme or with inverted gradients. We show that both approaches can match SAC's sample efficiency performance without the need of entropy maximization, we then propose a simple non-uniform sampling method for selecting transitions from the replay buffer during training. Extensive experimental results demonstrate that our proposed sampling scheme leads to state of the art sample efficiency on challenging continuous control tasks. We combine all of our findings into one simple algorithm, which we call Streamlined Off Policy with Emphasizing Recent Experience, for which we provide robust public-domain code. 
\end{abstract}

\section{Introduction}

Off-policy Deep Reinforcement Learning (RL) algorithms aim to improve sample efficiency by reusing past experience. 
Recently a number of new off-policy Deep RL algorithms have been proposed for control tasks with continuous state and action spaces, including Deep Deterministic Policy Gradient (DDPG)  and Twin Delayed DDPG (TD3) \citep{lillicrap2015ddpg,fujimoto2018td3}. TD3, which introduced clipped double-Q learning, delayed policy updates and target policy smoothing, has been shown to be significantly more sample efficient than popular on-policy methods for a wide range of MuJoCo benchmarks. 

The field of Deep Reinforcement Learning (DRL) has also recently seen a surge in the popularity of maximum entropy RL algorithms. 
In particular, Soft Actor-Critic (SAC), which combines off-policy learning with maximum-entropy RL, not only has many attractive theoretical properties, but can also give superior performance on a wide-range of MuJoCo environments, including on the high-dimensional environment Humanoid for which both DDPG and TD3 perform poorly \citep{haarnoja2018sac, haarnoja2018sacapps,langlois2019benchmarkingmodelbased}. SAC and TD3 have similar off-policy structures with clipped double-Q learning, but SAC also employs maximum entropy reinforcement learning. 

In this paper, we aim to develop off-policy DRL algorithms that not only provide state-of-the-art performance but are also simple and minimalistic.
We first seek to understand the primary contribution of the entropy term to the performance of maximum entropy algorithms.
For the MuJoCo benchmark, we demonstrate that when using the standard objective without entropy along with standard additive noise exploration, there is often insufficient exploration due to the bounded nature of the action spaces. Specifically, the outputs of the policy network are often way outside the bounds of the action space, so that they need to be squashed to fit within the action space. The squashing results in actions 
persistently taking on their maximal values, resulting in insufficient exploration.  
In contrast, the entropy term in the SAC objective forces the outputs to have sensible values, so that even with squashing, exploration is maintained. 
We conclude that, for the MuJoCo environments, the entropy term in the objective for Soft Actor-Critic principally addresses the bounded nature of the action spaces. 

With this insight, we propose the Streamlined Off Policy (SOP) algorithm, which is a minimalistic off-policy algorithm that includes a simple but crucial output normalization. 
The normalization addresses the bounded nature of the action spaces, allowing satisfactory exploration throughout training.
We also consider using inverting gradients (IG) \citep{hausknecht2015deep}
with the streamlined scheme, which we refer to as SOP\textunderscore IG.
Both approaches use the standard objective without the entropy term.  Our results show that SOP and SOP\textunderscore IG match the sample efficiency and robust performance of SAC, including on the  challenging Ant and Humanoid environments. 

Having matched SAC performance without using entropy maximization, we then seek to attain state-of-the-art performance by employing a non-uniform sampling method for selecting transitions from the replay buffer during training.
Priority Experience Replay (PER), a non-uniform sampling scheme, has been shown to significantly improve performance for the Atari games benchmark \citep{schaul2015prioritized}, but requires sophisticated data structure for efficient sampling.
Keeping with the theme of simplicity with the goal of meeting Occam's principle, we propose a novel and simple non-uniform sampling method for selecting transitions from the replay buffer during training.  Our method, called Emphasizing Recent Experience (ERE), samples more aggressively recent experience while not neglecting past experience. 
Unlike PER, ERE is only a few lines of code and does not rely on any sophisticated data structures. We show that when SOP, SOP\textunderscore IG, or SAC is combined with ERE, the resulting algorithm out-performs SAC and provides state of the art performance. For example, for Ant and Humanoid, SOP+ERE improves over SAC by $21\%$ and $24\%$, respectively, with one million samples.

The contributions of this paper are thus threefold. First, we uncover the primary contribution of the entropy term of maximum entropy RL algorithms for the MuJoCo environments. Second, we propose a streamlined algorithm which does not employ entropy maximization but nevertheless matches the sampling efficiency and robust performance of SAC for the MuJoCo benchmarks. And third, we propose a simple non-uniform sampling scheme to achieve state-of-the art performance for the MuJoCo benchmarks. We provide public code for SOP+ERE for reproducibility \footnote{\url{https://github.com/AutumnWu/Streamlined-Off-Policy-Learning}}.

\section{Preliminaries}

We represent an environment as a Markov Decision Process (MDP) which is defined by the tuple $(\mathcal{S}, \mathcal{A}, r, p, \gamma)$, where $\mathcal{S}$ and $\mathcal{A}$ are continuous multi-dimensional state and action spaces, $r(s,a)$ is a bounded reward function, $p(s'|s,a)$ is a transition function, and $\gamma$ is the discount factor. Let $s(t)$ and $a(t)$ respectively denote the state of the environment and the action chosen at time $t$. Let $\pi = \pi(a|s), \; s \in \gS, a \in \gA$ denote the policy. We further denote $K$ for the dimension of the action space, and write $a_k$ for the $k$th component of an action $a \in \gA$, that is, $a = (a_1,\ldots,a_K)$.

The expected discounted return for policy $\pi$ beginning in state $s$ is given by:
\begin{equation}
\label{standard_return}
V_\pi(s)=\E_\pi[\sum_{t=0}^{\infty} \gamma^t r(s(t),a(t)) | s(0)=s]
\end{equation}
Standard MDP and RL problem formulations seek to maximize $V_\pi(s)$ over policies $\pi$. For finite state and action spaces, under suitable conditions for continuous state and action spaces, there exists an optimal policy that is deterministic \citep{puterman2014markov, bertsekas1996neuro}. In RL with unknown environment, exploration is required to learn a suitable policy. 

In DRL with continuous action spaces, typically the policy is modeled by a parameterized policy network which takes as input a state $s$ and outputs a value $\mu(s; \theta)$, where $\theta$ represents the current parameters of the policy network 
\citep{schulman2015trpo,schulman2017ppo,vuong2018spu,lillicrap2015ddpg, fujimoto2018td3}. During training, typically additive random noise is added for exploration, so that the actual action taken when in state $s$ takes the form $a = \mu(s; \theta) + \epsilon$ where $\epsilon$ is a $K$-dimensional Gaussian random vector with each component having zero mean and variance $\sigma$. During testing, $\epsilon$ is set to zero.

\subsection{Maximum Entropy Reinforcement Learning}

Maximum entropy reinforcement learning takes a different approach than Equation (\ref{standard_return}) by optimizing policies to maximize both the expected return and the expected entropy of the policy \citep{ziebart2008maximum,ziebart2010modeling,todorov2008general,rawlik2012stochastic,levine2013guided, levine2016end,nachum2017bridging,haarnoja2017reinforcement,haarnoja2018sac,haarnoja2018sacapps}.

In particular, the maximum entropy RL objective is:
\begin{align*}
V_\pi(s)=
\sum_{t=0}^{\infty} \gamma^t  \E_\pi[ r(s(t),a(t)) \\+ \lambda H(\pi(\cdot|s(t))) | s(0)=s]
\end{align*}
where $H(\pi(\cdot|s))$ is the entropy of the policy when in state $s$, and
the temperature parameter $\lambda$ determines the relative importance of the entropy term against the reward. For maximum entropy DRL, when given state $s$ the policy network will typically output a $K$-dimensional vector $\sigma(s; \theta)$ in addition to the vector $\mu(s;\theta)$.  The action selected when in state $s$ is then modeled as $\mu(s;\theta) + \epsilon$ where $\epsilon \sim N(0,\sigma(s; \theta))$.

Maximum entropy RL has been touted to have a number of conceptual and practical advantages for DRL \citep{haarnoja2018sac,haarnoja2018sacapps}. For example, it has been argued that the policy is incentivized to explore more widely, while giving up on clearly unpromising avenues.
It has also been argued that the policy can capture multiple modes of near-optimal behavior, that is, in problem settings where multiple actions seem equally attractive, the policy will commit equal
probability mass to those actions. In this paper, we show for the MuJoCo benchmarks that the standard additive noise exploration suffices 
and can achieve the same performance as maximum entropy RL. 

\section{The Squashing Exploration Problem}

\subsection{Bounded Action Spaces}

\begin{figure*}[h!tb]
\centering
\begin{subfigure}{0.3\textwidth}
	\centering
	\includegraphics[width=0.95\linewidth]{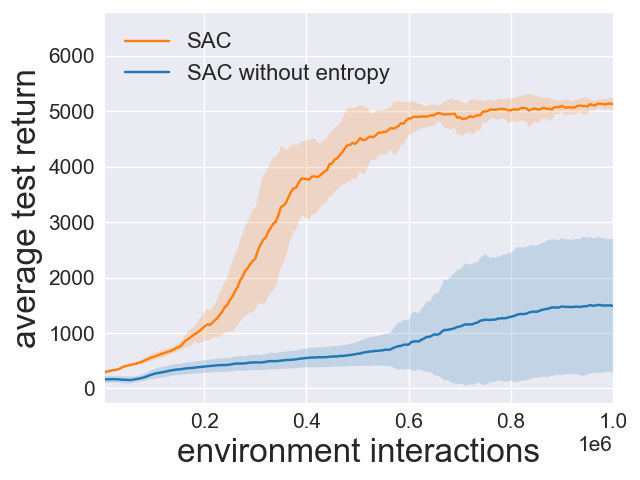}
	\caption{Humanoid-v2}
	\label{fig:humanoid_traincurve}
\end{subfigure}
\begin{subfigure}{0.3\textwidth}
	\centering
	\includegraphics[width=0.95\linewidth]{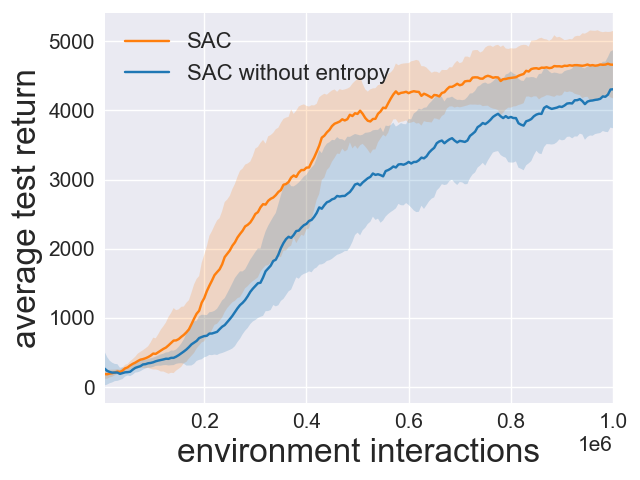}
	\caption{Walker2d-v2}
    \label{fig:walker2d_traincurve}
\end{subfigure}
\caption{SAC performance with and without entropy maximization}
\label{fig:SAC_training_curves}
\end{figure*}

Continuous environments typically have bounded action spaces, that is, along each action dimension $k$ there is a minimum possible action value $a_k^{\min}$ and a maximum possible action value $a_k^{\max}$. When selecting an action, the action needs to be selected within these bounds before the action can be taken. DRL algorithms often handle this by squashing the action so that it fits within the bounds. For example, if along any one dimension the value $\mu(s;\theta) + \epsilon$ exceeds $a^{\max}$, the action is set (clipped) to $a^{\max}$. Alternatively, a smooth form of squashing can be employed. For example, suppose $a_k^{\min} = - M$ and  $a_k^{\max} = + M$ for some positive number $M$, then a smooth form of squashing could use $a = M \tanh(\mu(s;\theta) + \epsilon )$ in which $\tanh()$ is being applied to each component of the $K$-dimensional vector. 
DDPG \citep{hou2017ddpgper} and TD3 \citep{fujimoto2018td3} use clipping, and SAC \citep{haarnoja2018sac,haarnoja2018sacapps} uses smooth squashing with the $\tanh()$ function. For concreteness, henceforth we will assume that smooth squashing with the $\tanh()$ is employed.

We note that an environment may actually allow the agent to input actions that are outside the bounds. In this case, the environment will typically first clip the actions internally before passing them on to the ``actual'' environment \citep{fujita2018clipped}. 

We now make a simple but crucial observation: squashing actions to fit into a bounded action space can have a disastrous effect on additive-noise exploration strategies. To see this, let the output of the policy network be $\mu(s) = (\mu_1(s),\ldots,\mu_K(s))$. Consider an action taken along one dimension $k$, and suppose $\mu_k(s) >> 1$ and $|\epsilon_k|$ is relatively small  compared to $\mu_k(s)$.
Then the action $a_k = M \tanh(\mu_k(s) + \epsilon_k )$ will be very close (essentially equal) to $M$. If the condition $\mu_k(s) >> 1$ persists over many consecutive states, then $a_k$ will remain close to 1 for all these states, and consequently there will be essentially no exploration along the $k$th dimension. We will refer to this problem as the {\em squashing exploration problem}. 
We will argue that algorithms using the standard objective (Equation \ref{standard_return}) with additive noise exploration can be greatly impaired by squashing exploration.   

\subsection{How Does Entropy Maximization Help for the MuJoCo Environments?}
\label{SAC_with_without_entropy}

\begin{figure*}[h!tb]
\centering
\begin{subfigure}{\textwidth}
    \centering
	\includegraphics[width=0.235\linewidth]{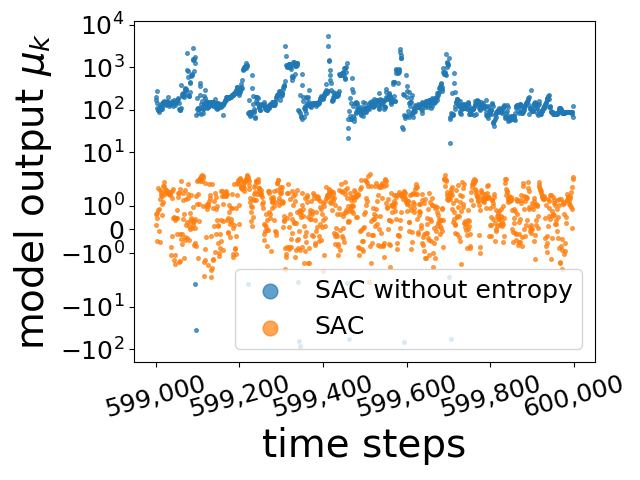}
	\label{fig:humanoid_muk_1}
	\includegraphics[width=0.235\linewidth]{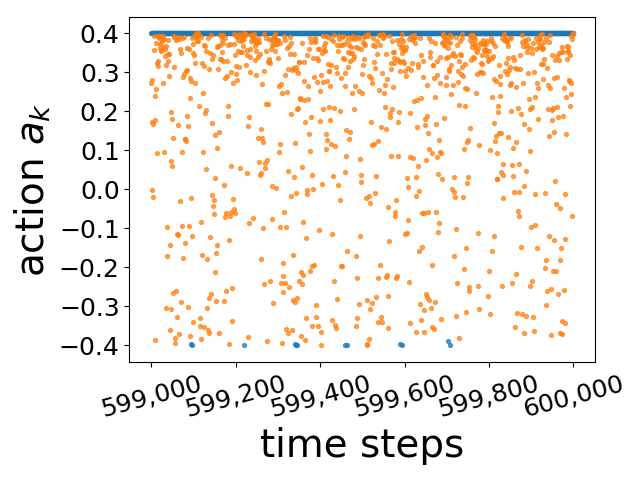}
    \label{fig:humanoid_muk_2}
	\includegraphics[width=0.235\linewidth]{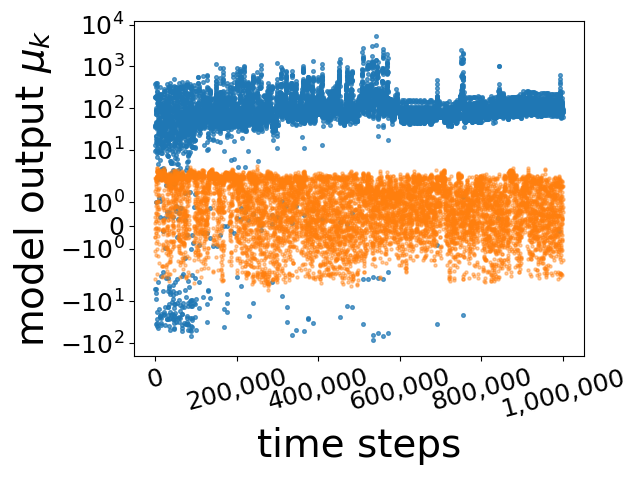}
    \label{fig:humanoid_muk_3}
	\includegraphics[width=0.235\linewidth]{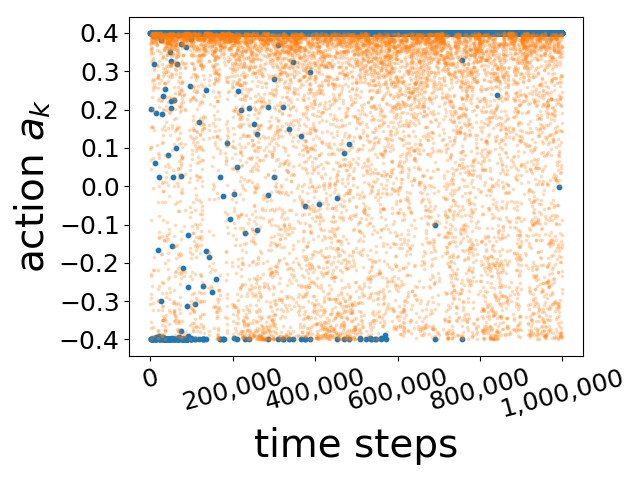}
    \label{fig:humanoid_muk_4}
    \caption{Humanoid-v2}
\end{subfigure}
\begin{subfigure}{\textwidth}
	\centering
	\includegraphics[width=0.235\linewidth]{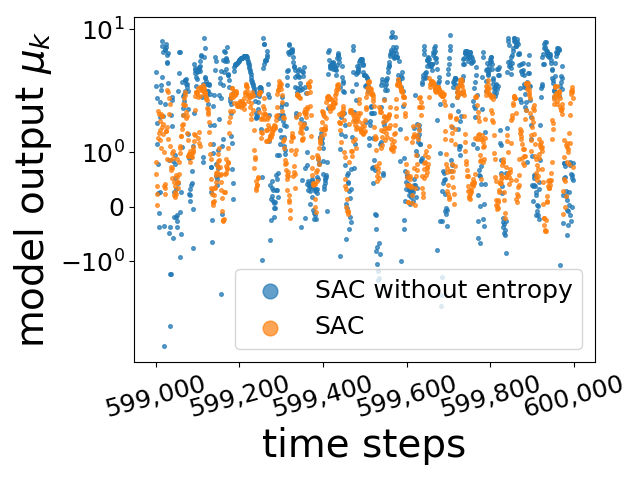}
    \label{fig:walker_muk_1}
	\includegraphics[width=0.235\linewidth]{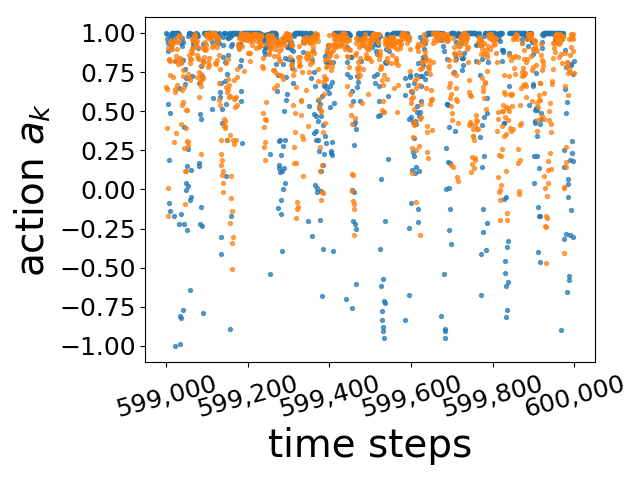}
    \label{fig:walker_muk_2}
	\includegraphics[width=0.235\linewidth]{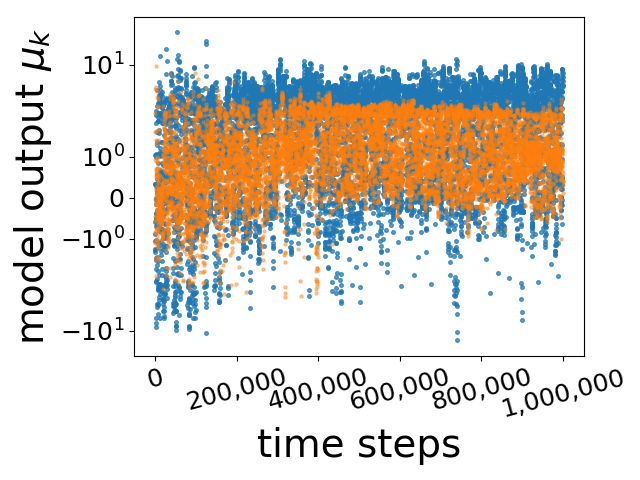}
    \label{fig:walker_muk_3}
	\includegraphics[width=0.235\linewidth]{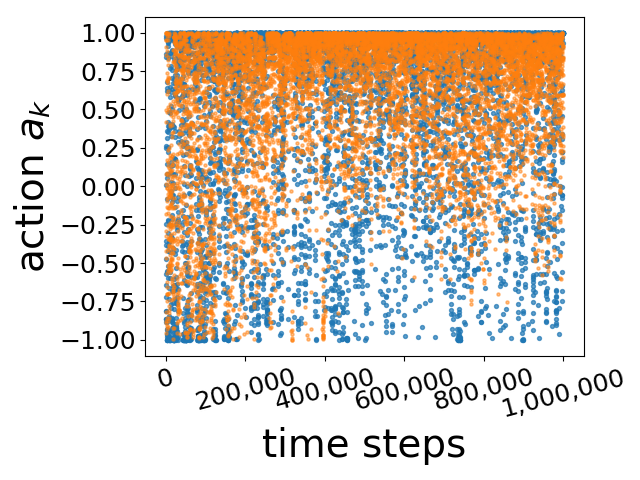}
    \label{fig:walker_muk_4}
    \caption{Walker2d-v2}
\end{subfigure}
\caption{$\mu_k$ and $a_k$ values from SAC and SAC without entropy maximization. See section 3.2 for a discussion.}
\label{fig:SAC_actions}
\end{figure*}

SAC is a maximum-entropy off-policy DRL algorithm which provides good performance across all of the MuJoCo benchmark environments. To the best of our knowledge, it currently provides state of the art performance for the MuJoCo benchmark. 
In this section, we argue that the principal contribution of the entropy term in the SAC objective is to resolve the squashing exploration problem, thereby 
maintaining sufficient exploration when facing bounded action spaces. To argue this, we consider two DRL algorithms: SAC with adaptive temperature \citep{haarnoja2018sacapps}, and SAC with entropy removed altogether (temperature set to zero) but everything else the same. We refer to them as {\em SAC} and as {\em SAC without entropy}. For SAC without entropy, for exploration we use additive zero-mean Gaussian noise with $\sigma$ fixed at $0.3$. Both algorithms use $\tanh$ squashing. We compare these two algorithms on two MuJoCo environments: Humanoid-v2 and Walker-v2. 

Figure \ref{fig:SAC_training_curves} shows the performance of the two algorithms with 10 seeds. For Humanoid, SAC performs much better than SAC without entropy. However, for Walker, SAC without entropy performs nearly as well as SAC, implying maximum entropy RL is not as critical for this environment. 

To understand why entropy maximization is important for one environment but less so for another, we examine the actions selected when training these two algorithms. Humanoid and Walker have action dimensions $K=17$ and $K=6$, respectively. Here we show representative results for one dimension for both environments. 
The top and bottom rows of Figure \ref{fig:SAC_actions} shows results for Humanoid and Walker, respectively. The first column shows the $\mu_k$ values for an interval of 1,000 consecutive time steps, namely, for time steps 599,000 to 600,000. The second column shows the actual action values passed to the environment for these time steps. The third and fourth columns show a concatenation of 10 such intervals of 1000 time steps, with each interval coming from a larger interval of 100,000 time steps. 

The top and bottom rows of Figure \ref{fig:SAC_actions} are strikingly different. For Humanoid using SAC with entropy, the $|\mu_k|$ values are small, mostly in the range [-1.5,1.5], and fluctuate significantly. This allows the action values to also fluctuate significantly, providing exploration in the action space. On the other hand, for SAC without entropy the $|\mu_k|$ values are typically huge, most of which are well outside the interval [-10,10]. This causes the actions $a_k$ to be persistently clustered at either $M$ or -$M$, leading to essentially no exploration along that dimension.  
For Walker, we see that for both algorithms, the $\mu_k$ values are sensible, mostly in the range [-1,1] and therefore the actions chosen by both algorithms exhibit exploration.

In conclusion, the principal benefit of maximum entropy RL in SAC for the MuJoCo environments is that it resolves the squashing exploration problem. 
For some environments (such as Walker), the outputs of the policy network take on sensible values, so that sufficient exploration is maintained and overall good performance is achieved without the need for entropy maximization. For other environments (such as Humanoid), entropy maximization is needed to reduce the magnitudes of the outputs %of the policy network 
so that exploration is maintained and overall good performance is achieved. 

\section{Matching SOTA Performance without Entropy Maximization}
In this paper we examine two approaches for matching SAC performance without using entropy maximization.

\begin{figure*}[h!tb]
\centering
\begin{subfigure}{0.3\textwidth}
	\centering
	\includegraphics[width=0.95\linewidth]{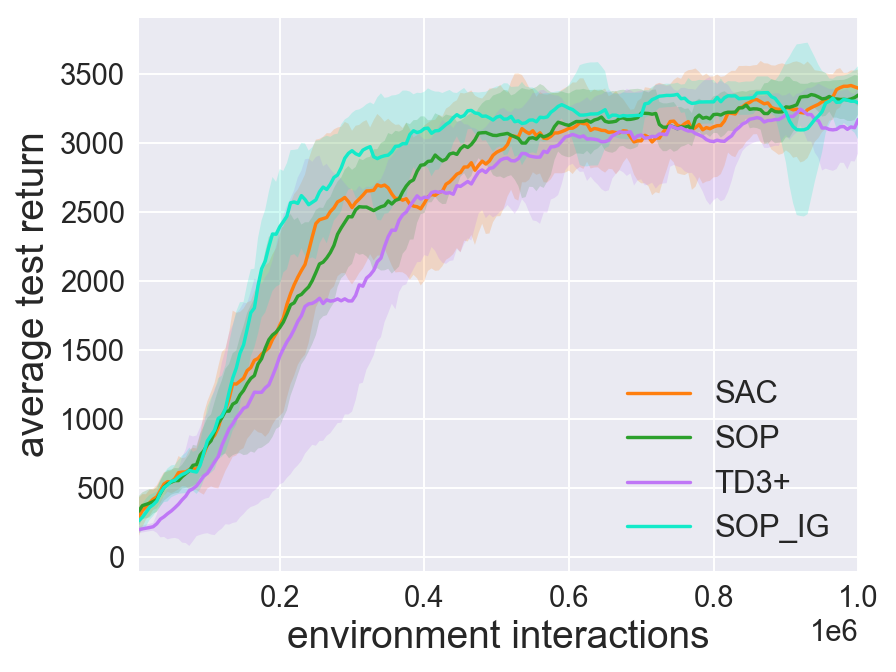}
	\caption{Hopper-v2}
	\label{fig:sop-sac-hopper}
\end{subfigure}
\begin{subfigure}{0.3\textwidth}
	\centering
	\includegraphics[width=0.95\linewidth]{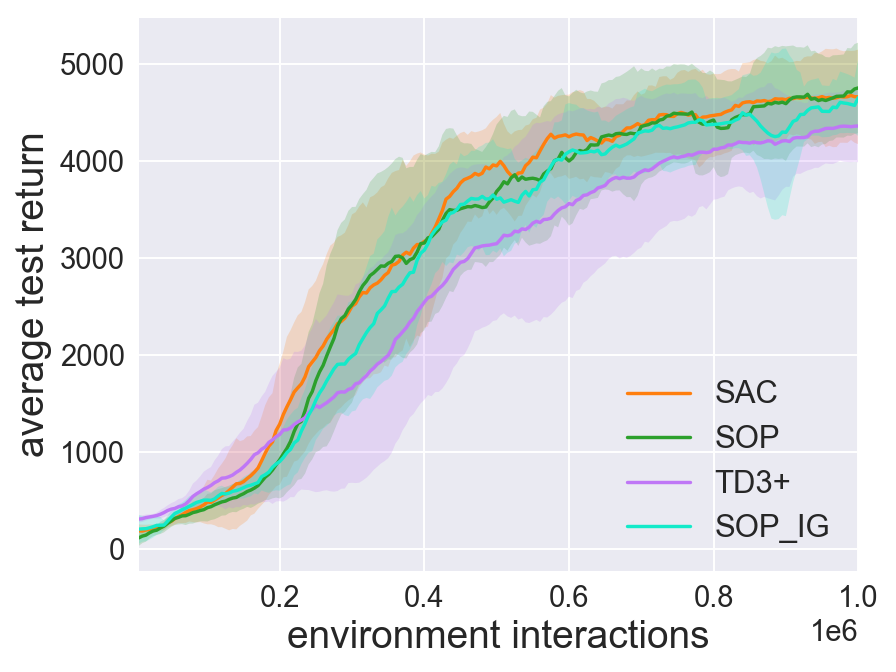}
	\caption{Walker2d-v2}
    \label{fig:sop-sac-walker2d}
\end{subfigure}
\begin{subfigure}{0.3\textwidth}
	\centering
	\includegraphics[width=0.95\linewidth]{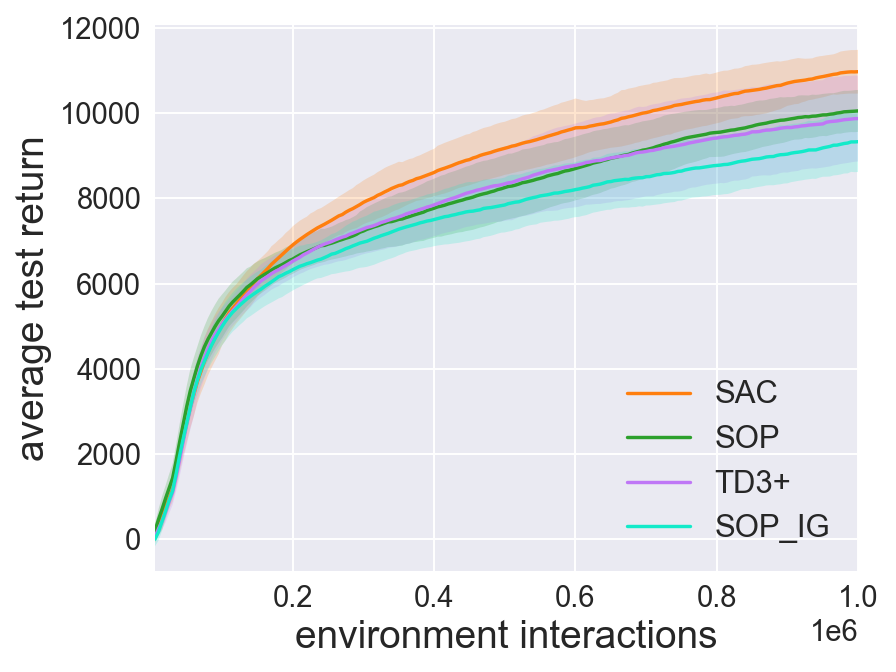}
	\caption{HalfCheetah-v2}
	\label{fig:sop-sac-halfcheetah}
\end{subfigure}
\begin{subfigure}{0.3\textwidth}
	\centering
	\includegraphics[width=0.95\linewidth]{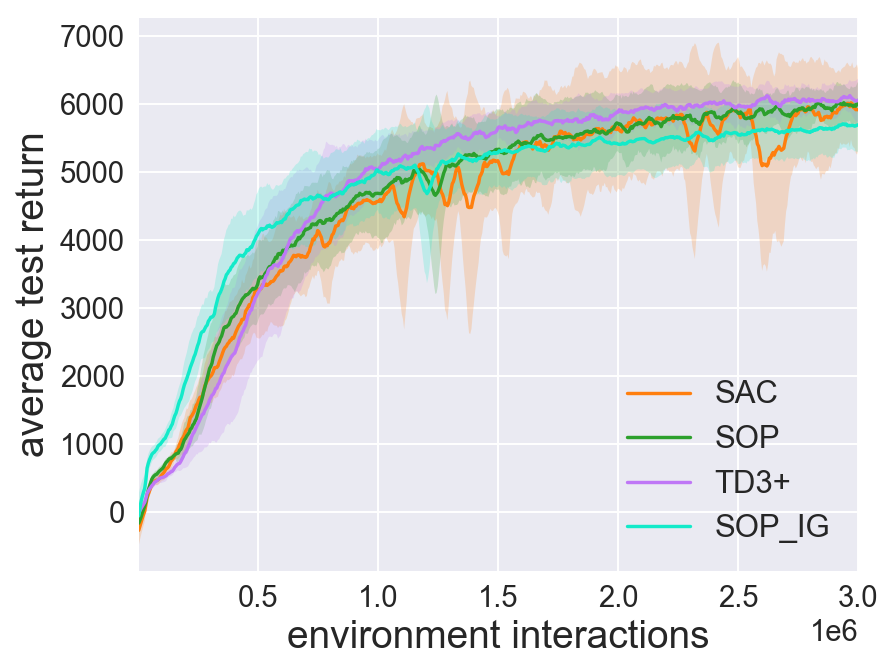}
	\caption{Ant-v2}
	\label{fig:sop-sac-ant}
\end{subfigure}
\begin{subfigure}{0.3\textwidth}
	\centering
	\includegraphics[width=0.95\linewidth]{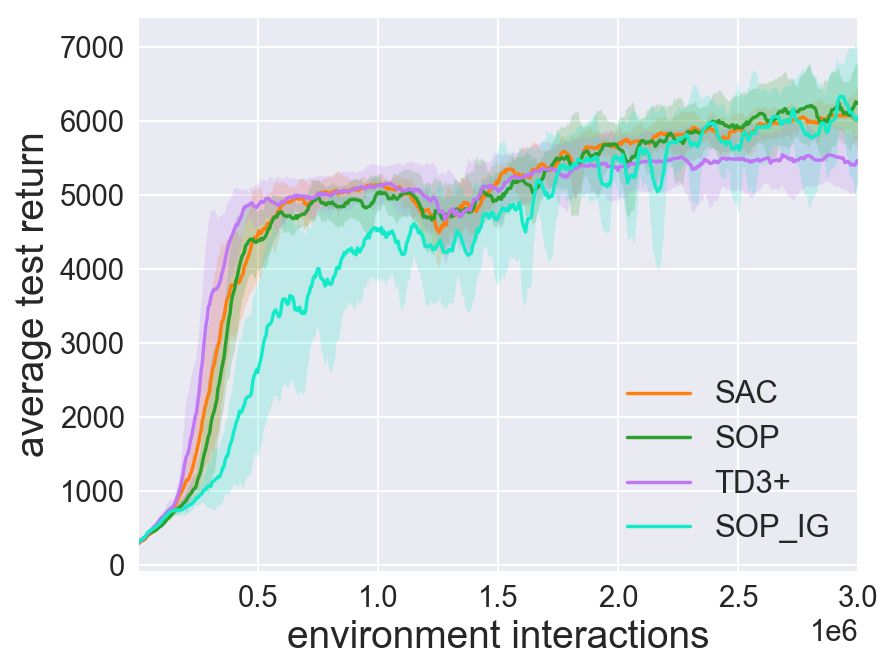}
	\caption{Humanoid-v2}
	\label{fig:sop-sac-humanoid}
\end{subfigure}
\caption{Streamlined Off-Policy (SOP) versus SAC, SOP\textunderscore IG and TD3}
\label{fig:SOP-SAC}
\end{figure*}

\subsection{Output Normalization}
As we observed in the previous section, in some environments the policy network output values $|\mu_k|$, $k=1,\ldots,K$ can become persistently huge, which leads to insufficient exploration due to the squashing. We propose a  simple solution of normalizing the outputs of the policy network when they collectively (across the action dimensions) become too large.
To this end, let $\mu = (\mu_1,\dots,\mu_K)$ be the output of the original policy network, and let 
$G= \sum_k |\mu_k|/K$. The $G$ is simply the average of the magnitudes of the components of $\mu$. 
The normalization procedure is as follows. 
If $G>1$, then we reset $\mu_k \leftarrow \mu_k/G$ for all $k=1,\ldots,K$; otherwise, we leave $\mu$ unchanged. With this simple normalization, we are assured that the average of the normalized magnitudes is never greater than one. 

Our Streamlined Off Policy (SOP) algorithm is described in Algorithm \ref{alg:sop}. The algorithm is essentially TD3 minus the delayed policy updates and the target policy parameters but with the addition of the normalization described above. SOP also uses $\tanh$ squashing instead of clipping, since $\tanh$ gives somewhat better performance in our experiments. 
The SOP algorithm is ``streamlined'' as it has no entropy terms, temperature adaptation, target policy parameters or delayed policy updates.
\subsection{Inverting Gradients}
In our experiments, we also consider using SOP but replacing the output normalization with the IG scheme \citep{hausknecht2015deep}. In this scheme, when gradients suggest increasing the action magnitudes, gradients are down scaled if actions are within the boundaries, and inverted if otherwise. 
More specifically, let $p$ be the output of the last layer of the policy network, let $p_{\min}$ and $p_{\max}$ be the action boundaries. The IG approach can be summarized as follows \citep{hausknecht2015deep}: 
\begin{equation}
  \nabla_{p} = \nabla_{p} \cdot
  \begin{cases} 
  \frac{p_{\max} - p}{p_{\max} - p_{\min}} & \parbox[t]{0.3 \linewidth}{\RaggedRight if $\nabla_{p}$ suggests increasing $p$} \\ 
    \frac{p - p_{\min}}{p_{\max} - p_{\min}} &\text{otherwise}
  \end{cases}
\end{equation}
Where $\nabla_{p}$ is the gradient of the policy loss w.r.t to $p$. 
Although IG is not complicated, it is not as simple and straightforward as simply normalizing the outputs. We refer to SOP with IG as SOP\textunderscore IG. Implementation details can be found in the supplementary materials. 

\subsection{Experimental Results for SOP and SOP\textunderscore IG}

Figure \ref{fig:SOP-SAC} compares SAC (with temperature adaptation \citep{haarnoja2018sac,haarnoja2018sacapps}) with SOP, SOP\textunderscore IG, and TD3 plus the simple normalization (which we call TD3+) for five of the most challenging MuJoCo environments. Using the same baseline code, we train each of the algorithms with 10 seeds. Each algorithm performs five evaluation rollouts every 5000 environment steps. The solid curves correspond to the mean, and the shaded region to the standard deviation of the returns over seeds. Results show that SOP, the simplest of all the schemes,  performs as well or better than all other schemes. In particular, SAC and SOP have similar sample efficiency and robustness across all environments.
TD3+ has slightly weaker asymptotic performance for Walker and Humanoid. SOP\textunderscore IG initially learns slowly for Humanoid with high variance across random seeds, but gives similar asymptotic performance. 
These experiments confirm that the performance of SAC can be achieved without maximum entropy RL. 

\subsection{Ablation Study for SOP}
In this ablation study, we separately examine the importance of $(i)$ the normalization at the output of the policy network; $(ii)$ the double Q networks; $(iii)$ and randomization used in the line 9 of the SOP algorithm (that is, target policy smoothing \citep{fujimoto2018td3}). 

\begin{figure*}[htb]
\centering
\begin{subfigure}{0.3\textwidth}
	\centering
	\includegraphics[width=0.95\linewidth]{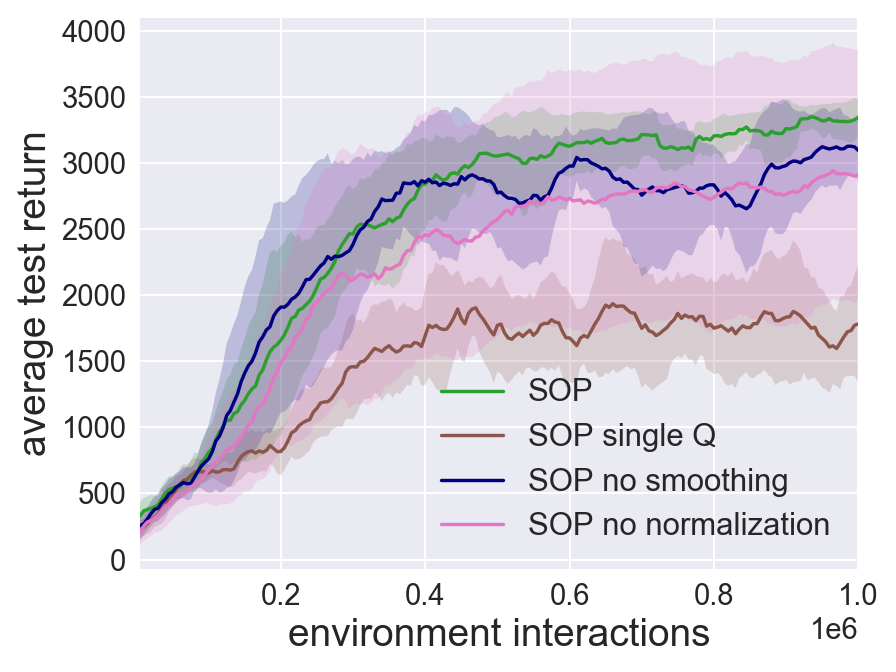}
	\caption{Hopper-v2}
	\label{fig:ablation-hopper}
\end{subfigure}
\begin{subfigure}{0.3\textwidth}
	\centering
	\includegraphics[width=0.95\linewidth]{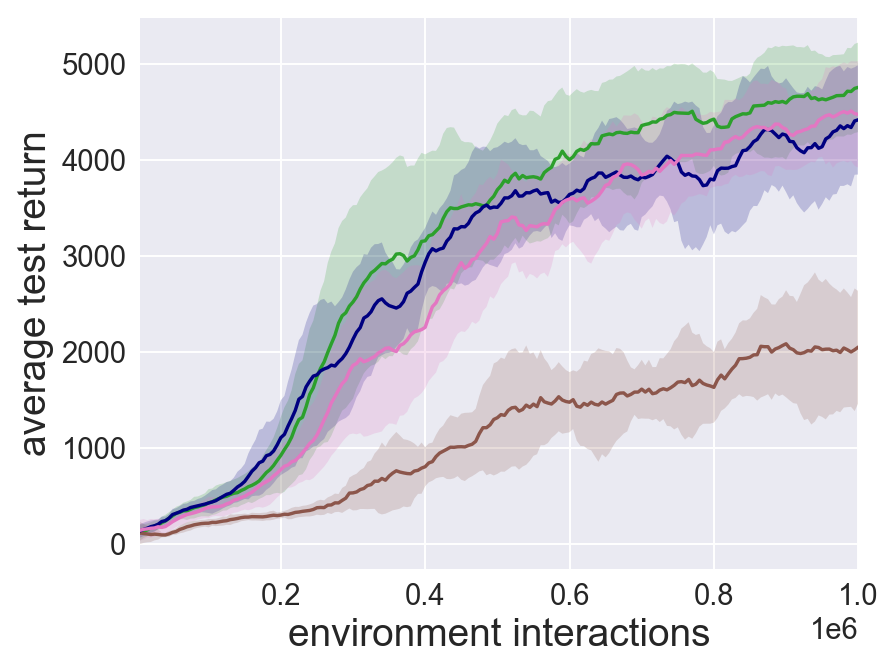}
	\caption{Walker2d-v2}
    \label{fig:ablation-walker2d}
\end{subfigure}
\begin{subfigure}{0.3\textwidth}
	\centering
	\includegraphics[width=0.95\linewidth]{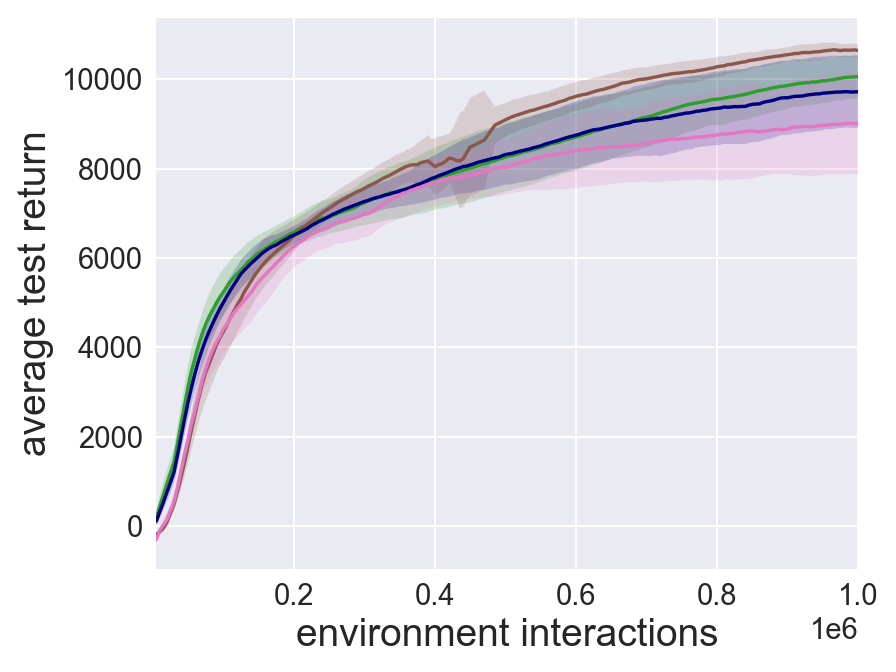}
	\caption{HalfCheetah-v2}
	\label{fig:ablation-halfcheetah}
\end{subfigure}
\begin{subfigure}{0.3\textwidth}
	\centering
	\includegraphics[width=0.95\linewidth]{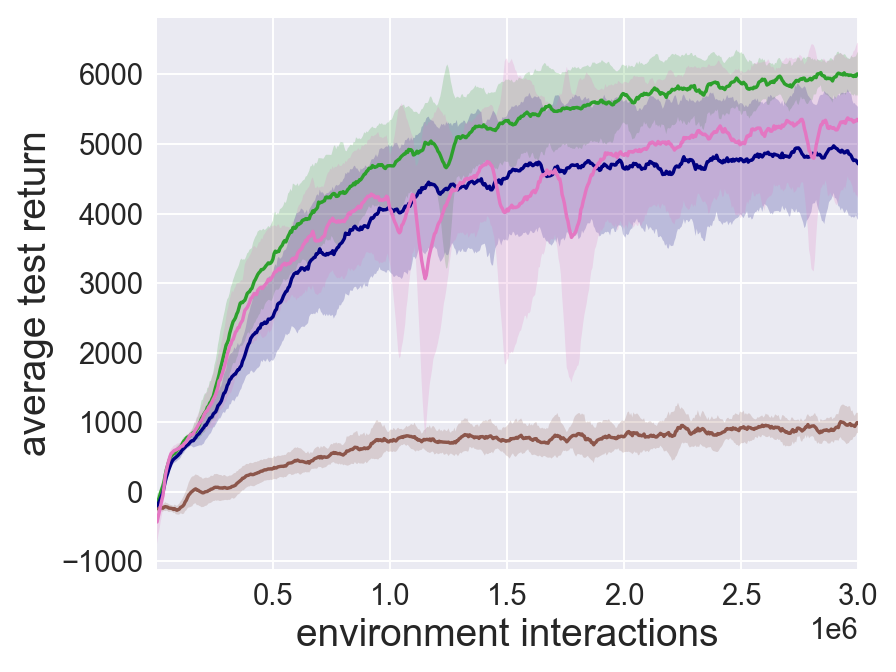}
	\caption{Ant-v2}
	\label{fig:ablation-ant}
\end{subfigure}
\begin{subfigure}{0.3\textwidth}
	\centering
	\includegraphics[width=0.95\linewidth]{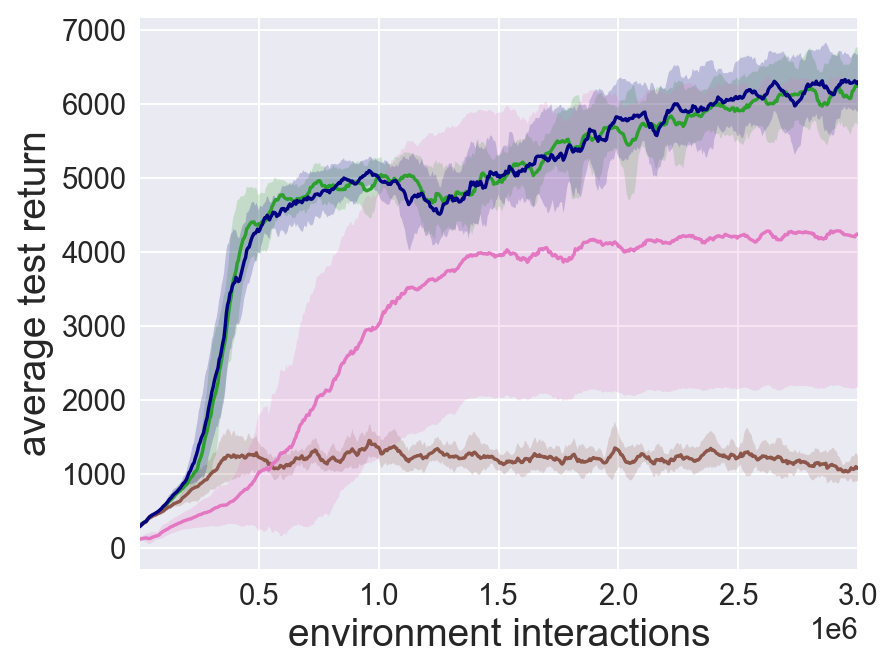}
	\caption{Humanoid-v2}
	\label{fig:ablation-humanoid}
\end{subfigure}
\caption{Ablation Study for SOP}
\label{fig:ablation}
\end{figure*}

Figure \ref{fig:ablation} shows the results for the five environments considered in this paper. In Figure \ref{fig:ablation}, ``no normalization'' is SOP without the normalization of the outputs of the policy network; ``single Q'' is SOP with one Q-network instead of two; and ``no smoothing'' is SOP without the randomness in line 8 of the algorithm. 

Figure \ref{fig:ablation} confirms that double Q-networks are critical for obtaining good performance \citep{van2016ddqn,fujimoto2018td3,haarnoja2018sac}. Figure \ref{fig:ablation} also shows that output normalization is critical. Without output normalization, performance fluctuates wildly, and average performance can decrease dramatically, particularly for Humanoid and HalfCheetah. Target policy smoothing improves performance by a relatively small amount.

In addition, to better understand whether the simple normalization term in SOP achieves a similar effect compared to explicitly maximizing entropy, we plot the entropy values for SOP and SAC throughout training for all environments. We found that SOP and SAC have very similar entropy values across training, while removing the entropy term from SAC makes the entropy value much lower. This indicates that the effect of the action normalization is very similar to maximizing entropy. The results can be found in the supplementary materials. 

\section{Non-Uniform Sampling}

\begin{figure*}[htb]
\centering
\begin{subfigure}{0.3\textwidth}
	\centering
	\includegraphics[width=0.95\linewidth]{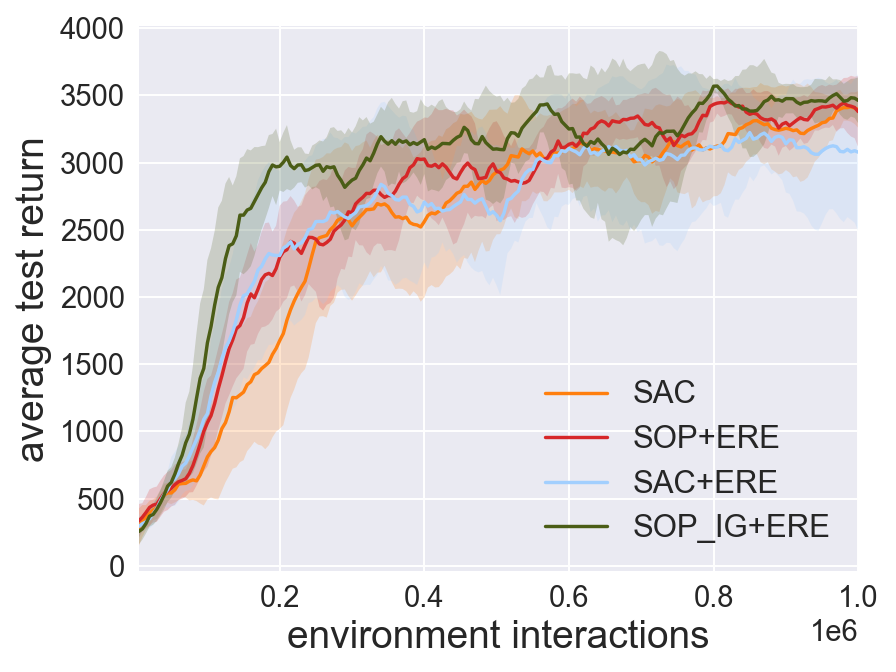}
	\caption{Hopper-v2}
\end{subfigure}
\begin{subfigure}{0.3\textwidth}
	\centering
	\includegraphics[width=0.95\linewidth]{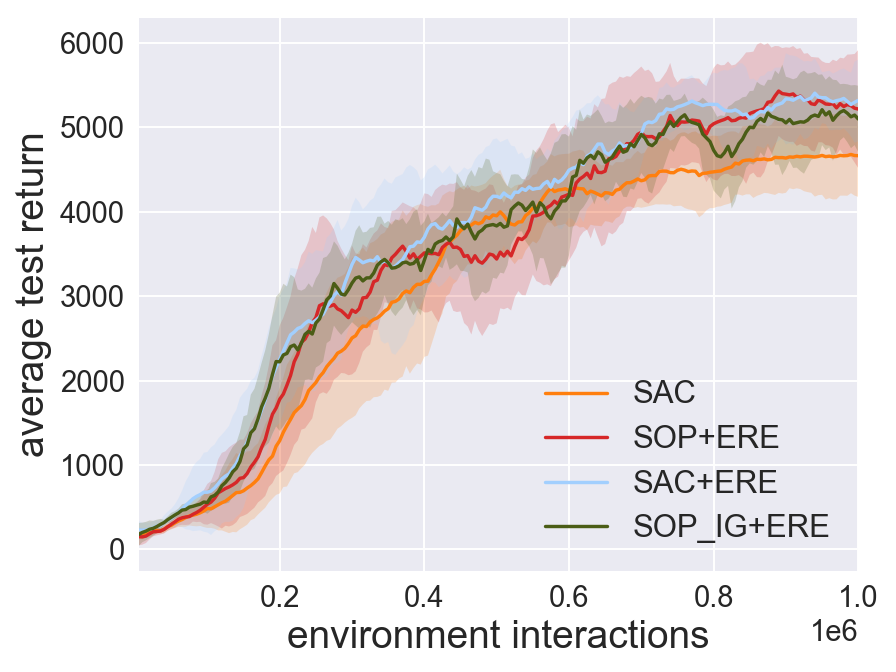}
	\caption{Walker2d-v2}
\end{subfigure}
\begin{subfigure}{0.3\textwidth}
	\centering
	\includegraphics[width=0.95\linewidth]{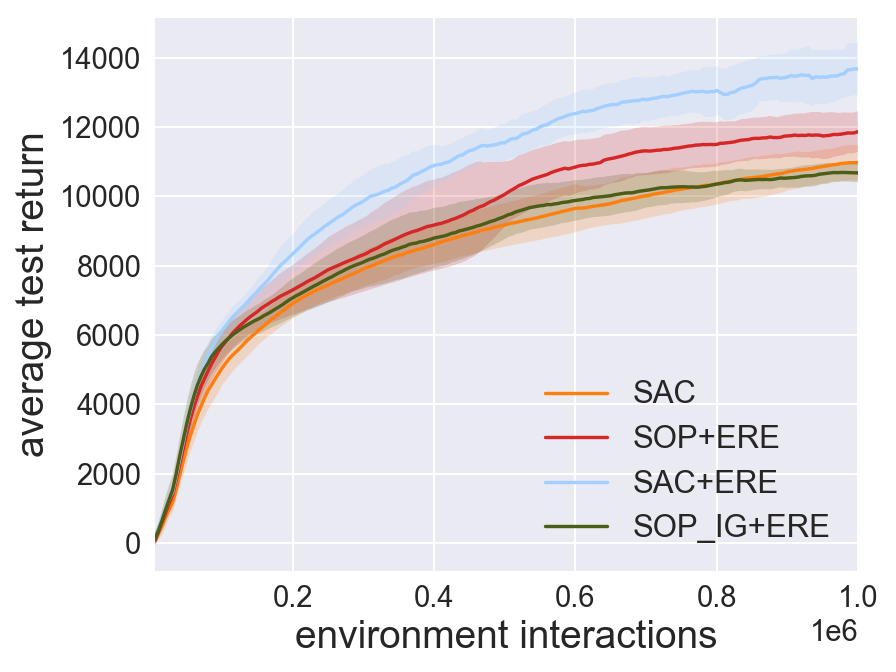}
	\caption{HalfCheetah-v2}
\end{subfigure}
\begin{subfigure}{0.3\textwidth}
	\centering
	\includegraphics[width=0.95\linewidth]{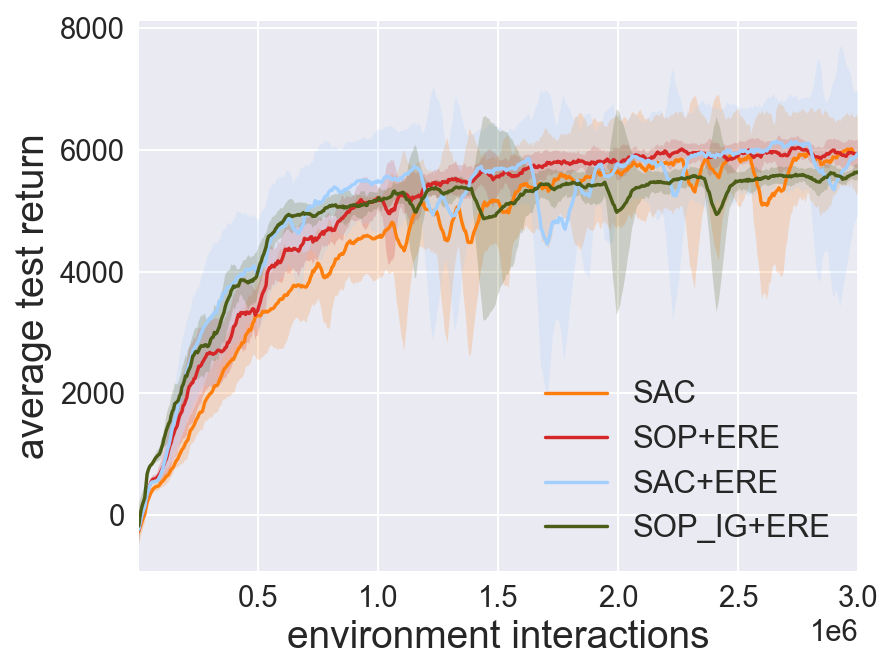}
	\caption{Ant-v2}
\end{subfigure}
\begin{subfigure}{0.3\textwidth}
	\centering
	\includegraphics[width=0.95\linewidth]{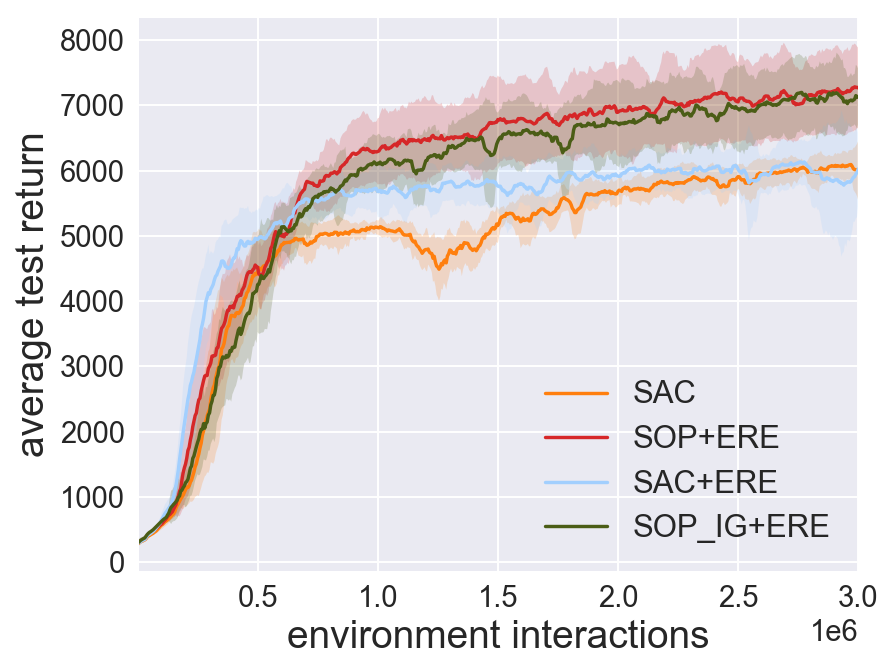}
	\caption{Humanoid-v2}
\end{subfigure}
\begin{subfigure}{0.3\textwidth}
	\centering
    \includegraphics[width=\linewidth]{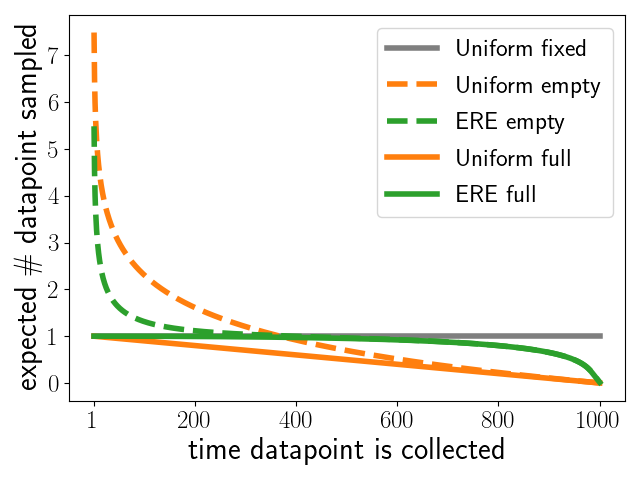}
    \caption{Uniform and ERE sampling}
    \label{fig:expected_update_done}
\end{subfigure}
\caption{(a) to (e) show the performance of SAC baseline, SOP+ERE, SAC+ERE, and SOP\textunderscore IG+ERE. (f) shows over a period of 1000 updates, the expected number of times the $t$th data point is sampled (with $\eta=0.996$). ERE allows new data to be sampled many times soon after being collected. 
}
\label{fig:sop-ere}
\end{figure*}

In the previous section we showed that SOP, SOP\textunderscore IG, and SAC all offer roughly equivalent sample-efficiency performance, with SOP being the simplest of the algorithms. 
We now show how a small change in the sampling scheme, which can be applied to any off-policy scheme (including SOP, SOP\textunderscore IG and SAC), can achieve state of the art performance for the MuJoCo benchmark. We call this non-uniform sampling scheme Emphasizing Recent Experience (ERE). ERE has 3 core features: $(i)$ It is a general method applicable to any off-policy algorithm; $(ii)$ It requires no special data structure, is very simple to implement, and has near-zero computational overhead; $(iii)$ It only introduces one additional important hyper-parameter.

The basic idea is that during the parameter update phase, the first mini-batch is sampled from the entire buffer, then for each subsequent mini-batch we gradually reduce our range of sampling to sample more from recent data. 
Specifically, assume that in the current update phase we are to make $1000$ mini-batch updates. Let $N$ be the max size of the buffer. Then for the $k^{th}$ update, we sample uniformly from the most recent $c_k$ data points, where $c_k = N \cdot \eta^{k}$ and $\eta \in (0,1]$ 
is a hyper-parameter that determines how much emphasis we put on recent data. 
$\eta=1$ is uniform sampling.
When $\eta < 1$, $c_k$ decreases as we perform each update. $\eta$ can be made to adapt to the learning speed of the agent so that we do not have to tune it for each environment.
The algorithmic and implementation details of such an adaptive scheme is given in the supplementary material.

The effect of such a sampling formulation is twofold. The first is recent data have a higher chance of being sampled. 
The second is that sampling is done in an ordered way: we first sample from all the data in the buffer, and gradually shrink the range of sampling to only sample from the most recent data. This scheme reduces the chance of over-writing parameter changes made by new data with parameter changes made by old data \citep{french1999catastrophic,mcclelland1995there, mccloskey1989catastrophic, ratcliff1990connectionist, robins1995catastrophic}. This allows us to quickly obtain information from recent data, and better approximate the value functions near recently-visited states, while still maintaining an acceptable approximation near states visited in the more distant past. 

What is the effect of replacing uniform sampling with ERE? First note if we uniformly sample several times from a fixed buffer (uniform fixed), where the buffer is filled, and no new data is coming in, then the expected number of times a data point has been sampled is the same for all data points. 

Now consider a scenario where we have a buffer of size 1000 (FIFO queue), we collect one data at a time, and then perform one update with mini-batch size of one. If we start with an empty buffer and sample uniformly (uniform empty), as data fills the buffer, each data point gets less and less chance of being sampled. Specifically, start from timestep 0, over a period of 1000 updates, the expected number of times the $t$th data (the data point collected at $t$th timestep) has been sampled is: $\frac{1}{t} + \frac{1}{t+1} + \dots + \frac{1}{1000}$. And if we start with a filled buffer and sample uniformly (uniform full), then the expected number of times the $t$th data has been sampled is $\sum_{t'=t}^{1000} \frac{1}{1000} = \frac{1000-t}{1000}$. 

Figure \ref{fig:expected_update_done} shows the expected number of times a data point has been sampled (at the end of 1000 updates) as a function of its position in the buffer. We see that when uniform sampling is used, older data are expected to get sampled much more than newer data, especially in the empty buffer case. 
This is undesirable because when the agent is improving and exploring new areas of the state space; new data points may contain more interesting information than the old ones, which have already been updated many times. 

When we apply the ERE scheme, we effectively skew the curve towards assigning higher expected number of samples for the newer data, allowing the newer data to be frequently sampled soon after being collected, which can accelerate the learning process. In Figure \ref{fig:expected_update_done} we can see that the curves for ERE (ERE empty and ERE full) are much closer to the horizontal line (Uniform fixed), compared to when uniform sampling is used. With ERE, at any point during training, we expect all data points currently in the buffer to have been sampled approximately the same number of times. Simply using a smaller buffer size will also allow recent data to be sampled more often, and can sometimes lead to a slightly faster learning speed in the early stage. However, it also tends to reduce the stability of learning, and damage long-term performance. 

Another simple method is to sample data according to an exponential scheme, where more recent data points are assigned exponentially higher probability of being sampled. In the supplementary materials, we provide further algorithmic detail and analysis on ERE, and compare ERE to the exponential sampling scheme, and show that ERE provides a stronger performance improvement. We also compare to another sampling scheme called Prioritized Experience Replay (PER) \citep{schaul2015prioritized}. PER assigns higher probability to data points that give a high absolute TD error when used for the Q update, then it applies an importance sampling weight according to the probability of sampling. Performance comparison can also be found in the supplementary materials. Results show that in the MuJoCo environments, PER can sometimes give a performance gain, but it is not as strong as ERE and the exponential scheme. 

\subsection{Experimental Results for ERE}
Figure \ref{fig:sop-ere} compares the performance of SAC (considered the baseline here), SAC+ERE, SOP+ERE, and SOP\textunderscore IG+ERE. ERE gives a significant boost to all three algorithms, surpassing SAC and achieving a new SOTA. Among the three algorithms, SOP+ERE gives the best performance for Ant and Humanoid (the two most challenging environments) and performance roughly equivalent to SAC+ERE and SOP\textunderscore IG+ERE for the other three environments. 

In particular, for Ant and Humanoid, SOP+ERE improves performance by 21\% and 24\% over SAC at 1 million samples, respectively. For Humanoid, at 3 million samples, SOP+ERE improves performance by 15\%.
In conclusion, SOP+ERE is not only a simple algorithm, but also exceeds state-of-the-art performance.

\begin{algorithm*}[htb]
    \caption{Streamlined Off-Policy}
	\label{alg:sop}
\begin{algorithmic}[1]
		\STATE Input: initial policy parameters $\theta$, Q-function parameters $\phi_1$, $\phi_2$, empty replay buffer $\mathcal{D}$
		\STATE Throughout the output of the policy network $\mu_{\theta}(s)$ is normalized if $ G > 1 $. (See Section 4.1.)
		\STATE Set target parameters equal to main parameters $\phi_{\text{targ}_i} \leftarrow \phi_i$ \;for i = 1, 2
		\REPEAT
		\STATE Generate an episode using actions $ a = M \text{tanh} (\mu_{\theta} (s) + \epsilon)$ where $\epsilon \sim \mathcal{N}(0,\sigma_1)$. 
		\FOR {$j$ in range(however many updates)}
		\STATE Randomly sample a batch of transitions, $B = \{ (s,a,r,s) \}$ from $\mathcal{D}$
		\STATE Compute targets for Q functions: \\
		\hskip1.5em $y_q (r,s') = r + \gamma \min_{i=1,2} Q_{\phi_{\text{targ}_i}}(s',  M \text{tanh} (\mu_{\theta} (s') + \delta )) \;\;\;\;\; \delta \sim \mathcal{N}(0, \sigma_2)$
		\STATE Update Q-functions by one step of gradient descent using \\
		\hskip1.5em $\nabla_{\phi_i} \frac{1}{|B|}\sum_{(s,a,r,s') \in B} \left( Q_{\phi_i}(s,a) - y_q(r,s') \right)^2 \text{for } i=1,2$
		\STATE Update policy by one step of gradient ascent using \\
		\hskip1.5em $\nabla_{\theta} \frac{1}{|B|}\sum_{s \in B}Q_{\phi_1}(s, M \tanh (\mu_{\theta}(s)))$
		\STATE Update target networks with \\
		\hskip1.5em $\phi_{\text{targ}_i} \leftarrow \rho \phi_{\text{targ}_i} + (1-\rho) \phi_i \;\;
		\text{for } i=1,2$
		\ENDFOR
		\UNTIL {Convergence}
\end{algorithmic}
\end{algorithm*}

\section{Related Work}

In recent years, there has been significant progress in improving the sample efficiency of DRL for continuous robotic locomotion tasks with off-policy algorithms \citep{lillicrap2015ddpg,fujimoto2018td3,haarnoja2018sac, haarnoja2018sacapps}. There is also a significant body of research on maximum entropy RL methods
\citep{ziebart2008maximum,ziebart2010modeling,todorov2008general,rawlik2012stochastic,levine2013guided, levine2016end,nachum2017bridging,haarnoja2017reinforcement,haarnoja2018sac,haarnoja2018sacapps}. \citet{ahmed2019understanding} very recently shed light on how entropy leads to a smoother optimization landscape. By taking clipping in the MuJoCo environments explicitly into account, \citet{fujita2018clipped} modified the policy gradient algorithm to reduce variance and provide superior performance among on-policy algorithms. \citet{eisenach2018marginal} extend the work of \citet{fujita2018clipped} for when an action may be direction. 
\citet{hausknecht2015deep} introduce Inverting Gradients, for which we provide expermintal results in this paper for the MuJoCo environments. \citet{chou2017improving} also explores DRL in the context of bounded action spaces. 
\citet{dalal2018safe} consider safe exploration in the context of constrained action spaces. 

Experience replay \citep{lin1992experiencereplay} is a simple yet powerful method for enhancing the performance of an off-policy DRL algorithm. Experience replay stores past experience in a replay buffer and reuses this past data when making updates. It achieved great successes in Deep Q-Networks (DQN) \citep{mnih2013dqn, mnih2015dqn}. 

Uniform sampling is the most common way to sample from a replay buffer. One of the most well-known alternatives is prioritized experience replay (PER) \citep{schaul2015prioritized}. PER uses the absolute TD-error of a data point as the measure for priority, and data points with higher priority will have a higher chance of being sampled. This method has been tested on DQN \citep{mnih2015dqn} and double DQN (DDQN) \citep{van2016ddqn} with significant improvement and applied successfully in other algorithms \citep{wang2015dueling,schulze2018vizdoom, hessel2018rainbow, hou2017ddpgper} and can be implemented in a distributed manner \citep{horgan2018distributed}.

When new data points lead to large TD errors in the Q update, PER will also assign high sampling probability to newer data points. However, PER has a different effect compared to ERE. PER tries to fit well on both old and new data points. While for ERE, old data points are always considered less important than newer data points even if these old data points start to give a high TD error. A performance comparison of PER and ERE are given in the supplementary materials. 

There are other methods proposed to make better use of the replay buffer. The ACER algorithm has an on-policy part and an off-policy part, with a hyper-parameter controlling the ratio of off-policy to on-policy updates \citep{wang2016acer}. The RACER algorithm \citep{novati2018remember} selectively removes data points from the buffer, based on the degree of ``off-policyness,'' bringing improvement to DDPG \citep{lillicrap2015ddpg}, NAF \citep{gu2016naf} and PPO \citep{schulman2017ppo}. In \citet{de2015replaydatabase}, replay buffers of different sizes were tested, showing large buffer with data diversity can lead to better performance. Finally, with Hindsight Experience Replay\citep{andrychowicz2017her}, priority can be given to trajectories with lower density estimation \citep{zhao2019curiosity} to tackle multi-goal, sparse reward environments.

\section{Conclusion}
In this paper we first showed that the primary role of maximum entropy RL for the MuJoCo benchmark is to maintain satisfactory exploration in the presence of bounded action spaces. 
We then developed a new streamlined algorithm which does not employ entropy maximization but nevertheless matches the sampling efficiency and robust performance of SAC for the MuJoCo benchmarks. Finally, we combined our streamlined algorithm with a simple non-uniform sampling scheme to create a simple algorithm that achieves state-of-the art performance for the MuJoCo benchmark. 

\section*{Acknowledgements}
We would like to thank Yiming Zhang for insightful discussion of our work; Josh Achiam for his help with the OpenAI Spinup codebase. We would also like to thank the reviewers for their helpful and constructive comments.

\bibliography{icml2020}
\bibliographystyle{icml2020}

\icmltitlerunning{Supplementary Material for Striving for simplicity and performance in off-policy DRL}

%\begin{document}
\twocolumn[
\icmltitle{Supplementary Material for Striving for Simplicity and Performance in Off-Policy DRL:
Output Normalization and Non-Uniform Sampling}

\vskip 0.3in]

\section{Hyperparameters}
Table \ref{tab:sop-hyperparameter} shows hyperparameters used for SOP, SOP+ERE and SOP+PER. For adaptive SAC, we use our own PyTorch implementation for the comparisons. Our implementation uses the same hyperparameters as used in the original paper \citep{haarnoja2018sacapps}. Our implementation of SOP variants and adaptive SAC share most of the code base. For TD3, our implementation uses the same hyperparamters as used in the authors' implementation, which is different from the ones in the original paper \citep{fujimoto2018td3}. They claimed that the new set of hyperparamters can improve performance for TD3. We now discuss hyperparameter search for better clarity, fairness and reproducibility \citep{henderson2018matters, duan2016benchmarkingdrl, islam2017reproducibility}.

For the $\eta$ value in the ERE scheme, in our early experiments we tried the values (0.993, 0.994, 0.995, 0.996, 0.997, 0.998) on the Ant and found 0.995 to work well. This initial range of values was decided by computing the ERE sampling range for the oldest data. We found that for smaller values, the range would simply be too small. For the PER scheme, we did some informal preliminary search, then searched on Ant for $\beta_1$ in (0, 0.4, 0.6, 0.8), $\beta_2$ in (0, 0.4, 0.5, 0.6, 1), and learning rate in (1e-4, 2e-4, 3e-4, 5e-4, 8e-4, 1e-3), we decided to search these values because the original paper used $\beta_1 = 0.6$, $\beta_2=0.4$ and with reduced learning rate. For the exponential sampling scheme, we searched the $\lambda$ value in (3e-7, 1e-6, 3e-6, 5e-6, 1e-5, 3e-5, 5e-5, 1e-4) in Ant, this search range was decided by plotting out the probabilities of sampling, and then pick a set of values that are not too extreme. For $\sigma$ in SOP, in some of our early experiments with SAC, we accidentally found that $\sigma=0.3$ gives good performance for SAC without entropy and with Gaussian noise. We searched values (0.27, 0.28, 0.29, 0.3). For $\sigma$ values for TD3+, we searched values (0.1, 0.15, 0.2, 0.25, 0.3).

\begin{table*}[!htbp]
	\renewcommand{\arraystretch}{1.1}
	\centering
	\caption{SOP Hyperparameters}
	\label{tab:sop-hyperparameter}
	\vspace{1mm}
	\begin{tabular}{l l| l }
		\hline
		\multicolumn{2}{l|}{Parameter} &  Value\\
		\hline

		\multicolumn{2}{l|}{\it{Shared}}& \\
		& optimizer &Adam \citep{kingma2014adam}\\
		& learning rate & $3 \cdot 10^{-4}$\\
		& discount ($\gamma$) &  0.99\\
		& target smoothing coefficient ($\rho$)& 0.005\\
		& target update interval & 1\\
		& replay buffer size & $10^6$\\
		& number of hidden layers for all networks & 2\\
		& number of hidden units per layer & 256\\
		& mini-batch size & 256\\
		& nonlinearity & ReLU\\
		\hline
		
		\multicolumn{2}{l|}{\it{SAC adaptive}}& \\
		& entropy target & −dim($\gA$) (e.g., 6 for HalfCheetah-v2)\\
		\hline
		\multicolumn{2}{l|}{\it{SOP}}& \\
		& gaussian noise std $\sigma = \sigma_1 = \sigma_2$     &0.29\\
% 		&normalization constant $\beta$          &$1.2$ \\

        \hline
		\multicolumn{2}{l|}{\it{TD3}}& \\
		& gaussian noise std for data collection $\sigma$       &0.1 * action limit\\
		& guassian noise std for target policy smoothing $\Tilde{\sigma}$ &0.2\\
		\hline
		\multicolumn{2}{l|}{\it{TD3+}}& \\
		& gaussian noise std for data collection $\sigma$       &0.15\\
		& guassian noise std for target policy smoothing $\Tilde{\sigma}$ &0.2\\
		\hline
		
		\multicolumn{2}{l|}{\it{ERE}}& \\
		&ERE initial $\eta_0$          &$0.995$ \\
		\hline 
		\multicolumn{2}{l|}{\it{PER}}& \\
		%\hline 
		&PER $\beta_1$ ($\alpha$ in PER paper)            &$0.4$\\
		&PER $\beta_2$ ($\beta$ in PER paper)            &$0.4$\\
		\hline
		\multicolumn{2}{l|}{\it{EXP}}& \\
		%\hline 
		&Exponential $\lambda$            &$5e-06$\\
		\hline
		
	\end{tabular}
\end{table*}

\section{Entropy Value Comparison}
\begin{figure*}[h!tb]
\centering
\begin{subfigure}{0.3\textwidth}
	\centering
	\includegraphics[width=0.95\linewidth]{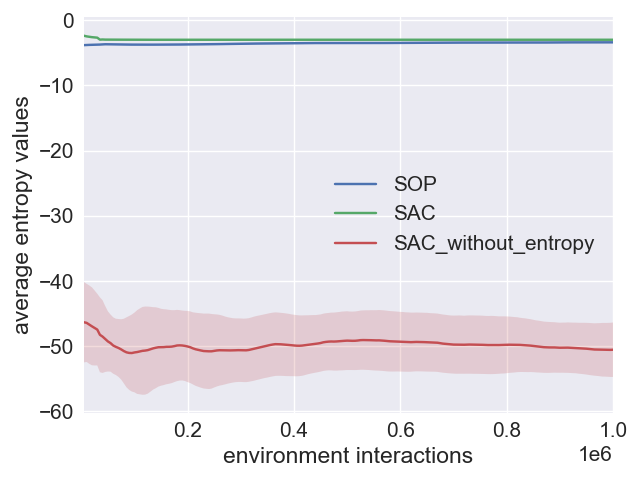}
	\caption{Hopper-v2}
	\label{fig:entropy-hopper}
\end{subfigure}
\begin{subfigure}{0.3\textwidth}
	\centering
	\includegraphics[width=0.95\linewidth]{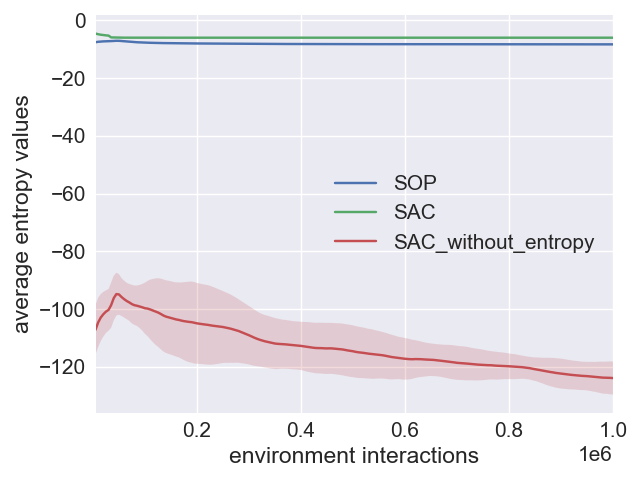}
	\caption{Walker2d-v2}
    \label{fig:entropy-walker2d}
\end{subfigure}
\begin{subfigure}{0.3\textwidth}
	\centering
	\includegraphics[width=0.95\linewidth]{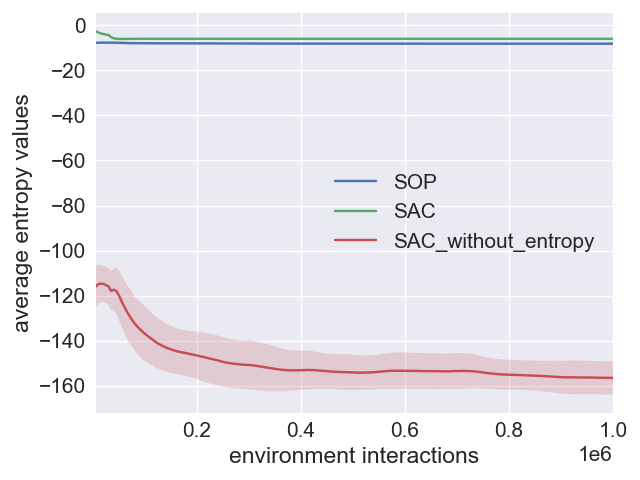}
	\caption{HalfCheetah-v2}
	\label{fig:entropy-halfcheetah}
\end{subfigure}
\begin{subfigure}{0.3\textwidth}
	\centering
	\includegraphics[width=0.95\linewidth]{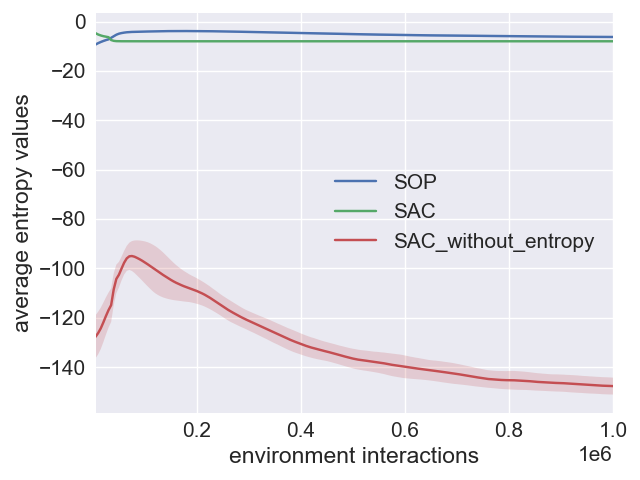}
	\caption{Ant-v2}
	\label{fig:entropy-ant}
\end{subfigure}
\begin{subfigure}{0.3\textwidth}
	\centering
	\includegraphics[width=0.95\linewidth]{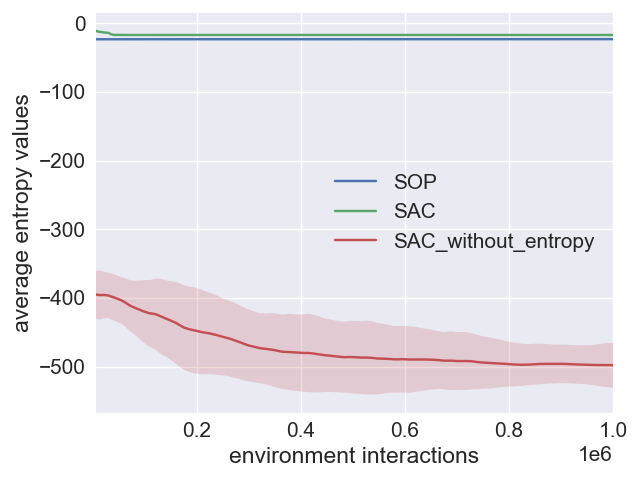}
	\caption{Humanoid-v2}
	\label{fig:entropy-humanoid}
\end{subfigure}
\caption{Entropy value comparison between SOP, SAC, and SAC without entropy maximization}
\label{fig:entropy-com}
\end{figure*}

To better understand whether the simple normalization term in SOP achieves a similar effect compared to explicitly maximizing entropy, we plot the entropy values for SOP and SAC throughout training for all environments. Figure \ref{fig:entropy-com} shows that the SOP and SAC policies have very similar entropy values across training, while removing the entropy term from SAC leads to a much lower entropy value. This indicates that the effect of the action normalization is very similar to maximizing entropy. 
 
%\clearpage
\section{ERE Pseudocode}
Our Streamlined Off Policy (SOP) with Emphasizing Recent Experience (ERE) algorithm is described in Algorithm \ref{alg:sop-ere}.

\begin{algorithm*}
	%\algtext*{End}
	\caption{SOP with Emphasizing Recent Experience}
	\label{alg:sop-ere}
	\begin{algorithmic}[1]
		\STATE Input: initial policy parameters $\theta$, Q-function parameters $\phi_1$, $\phi_2$, empty replay buffer $\mathcal{D}$ of size $N$, initial $\eta_0$, recent and max performance improvement $I_{recent} = I_{max} =0$.
		\STATE Set target parameters equal to main parameters $\phi_{\text{targ,i}} \leftarrow \phi_i$ \;for i = 1, 2
		%\MRepeat
		\REPEAT
		\STATE Generate an episode using actions $ a = M \text{tanh} (\mu_{\theta} (s) + \epsilon)$ where $\epsilon \sim \mathcal{N}(0,\sigma_1)$. 
		% 		\If{It's time to update}
		\STATE update $I_{recent}, I_{max}$ with training episode returns, let $K=$ length of episode
		\STATE compute $\eta = \eta_0 \cdot \frac{I_{recent}}{I_{max}} + (1-\frac{I_{recent}}{I_{max}})$ 
		\FOR {$j$ in range($K$)}
		\STATE Compute $c_k = N \cdot \eta^{k\frac{1000}{K}}$
		\STATE Sample a batch of transitions, $B = \{ (s,a,r,s) \}$ from most recent $c_k$ data in $\mathcal{D}$
		\STATE Compute targets for Q functions:
		
		\hskip1.5em  $y_q (r,s') = r + \gamma \min_{i=1,2} Q_{\phi_{\text{targ},i}}(s',  M \text{tanh} (\mu_{\theta} (s') + \delta )) \;\;\;\;\; \delta \sim \mathcal{N}(0, \sigma_2)$
%		\
		\STATE Update Q-functions by one step of gradient descent using \\
		\hskip1.5em  $\nabla_{\phi_i} \frac{1}{|B|}\sum_{(s,a,r,s') \in B} \left( Q_{\phi,i}(s,a) - y_q(r,s') \right)^2 \text{for } i=1,2$
		\STATE Update policy by one step of gradient ascent using \\
		\hskip1.5em  $\nabla_{\theta} \frac{1}{|B|}\sum_{s \in B}Q_{\phi,1}(s, M \tanh (\mu_{\theta}(s)))$
		\STATE Update target networks with \\
		\hskip1.5em  $\phi_{\text{targ, i}} \leftarrow \rho \phi_{\text{targ, i}} + (1-\rho) \phi_i \;\;
		\text{for } i=1,2$
		\ENDFOR
		% 		\EndIf
		\UNTIL{Convergence}
	\end{algorithmic}
\end{algorithm*}

%\clearpage

\section{Inverting Gradient Method}

In this section we discuss the details of the Inverting Gradient method. 

\citet{hausknecht2015deep} discussed three different methods for bounded parameter space learning: Zeroing Gradients, Squashing Gradients and Inverting Gradients, they analyzed and tested the three methods and found that Inverting Gradients method can achieve much stronger performance than the other two. In our implementation, we remove the tanh function from SOP and use Inverting Gradients instead to bound the actions. Let $p$ indicate the output of the last layer of the policy network. During exploration $p$ will be the mean of a normal distribution that we sample actions from, the IG approach can be summarized by the following equation \citep{hausknecht2015deep}: 

\begin{equation}
  \nabla_{p} = \nabla_{p} \cdot
  \begin{cases} 
  \frac{p_{\max} - p}{p_{\max} - p_{\min}} & \parbox[t]{0.3 \linewidth}{\RaggedRight if $\nabla_{p}$ suggests increasing $p$} \\ 
    \frac{p - p_{\min}}{p_{\max} - p_{\min}} &\text{otherwise}
  \end{cases}
\end{equation}

Where $\nabla_{p}$ is the gradient of the policy loss w.r.t to $p$. During a policy network update, we first backpropagate the gradients from the outputs of the Q network to the output of the policy network for each data point in the batch, we then compute the ratio $(p_{\max} - p) / (p_{\max} - p_{\min})$ or $(p_{\max} - p) / (p_{\max} - p_{\min})$ for each $p$ value (each action dimension), depending on the sign of the gradient. We then backpropagate from the output of the policy network to parameters of the policy network, and we modify the gradients in the policy network according to the ratios we computed. We made an efficient implementation and further discuss the computation efficiency of IG in the implementation details section. 

\section{SOP with Other Sampling Schemes}
We also investigate the effect of other interesting sampling schemes.
\subsection{SOP with Prioritized Experience Replay}
We also implement the proportional variant of Prioritized Experience Replay \citep{schaul2015prioritized} with SOP. 

Since SOP has two Q-networks, we redefine the absolute TD error $|\delta|$ of a transition $(s,a,r,s')$ to be the average absolute TD error in the Q network update:
\begin{equation}
|\delta| = \frac{1}{2} \sum_{l=1}^{2} |y_q(r,s') - Q_{\phi,l}(s, a)|
\end{equation}

Within the sum, the first term $y_q (r,s') = r + \gamma \min_{i=1,2} Q_{\phi_{\text{targ},i}}(s',  \text{tanh} (\mu_{\theta} (s') + \delta ))$, $\delta \sim \mathcal{N}(0, \sigma_2)$ is simply the target for the Q network, and the term $Q_{\theta,l}(s, a)$ is the current estimate of the $l^{th}$ Q network. 
For the $i^{th}$ data point, the definition of the priority value $p_i$ is $p_i = |\delta_i| +\epsilon$.
The probability of sampling a data point $P(i)$ is computed as:
\begin{equation}
    P(i) = \frac{p_i^{\beta_1}}{\sum_j p_j^{\beta_1}}
\end{equation}
where $\beta_1$ is a hyperparameter that controls how much the priority value affects the sampling probability, which is denoted by $\alpha$ in \citet{schaul2015prioritized}, but to avoid confusion with the $\alpha$ in SAC, we denote it as $\beta_1$. The importance sampling (IS) weight $w_i$ for a data point is computed as:
\begin{equation}
    w_i=(\frac{1}{N} \cdot \frac{1}{P(i)})^{\beta_2}
\end{equation}
where $\beta_2$ is denoted as $\beta$ in \citet{schaul2015prioritized}. 

Based on the SOP algorithm, we change the sampling method from uniform sampling to sampling using the probabilities $P(i)$, and for the Q updates we apply the IS weight $w_i$. This gives SOP with Prioritized Experience Replay (SOP+PER). 
We note that as compared with SOP+PER, ERE does not require a special data structure and has negligible extra cost, while PER uses a sum-tree structure with some additional computational cost.  
We also tried several variants of SOP+PER, but preliminary results show that it is unclear whether there is improvement in performance, so we kept the algorithm simple. 

\begin{figure*}[!ht]
\centering
\begin{subfigure}{0.3\textwidth}
	\centering
	\includegraphics[width=0.95\linewidth]{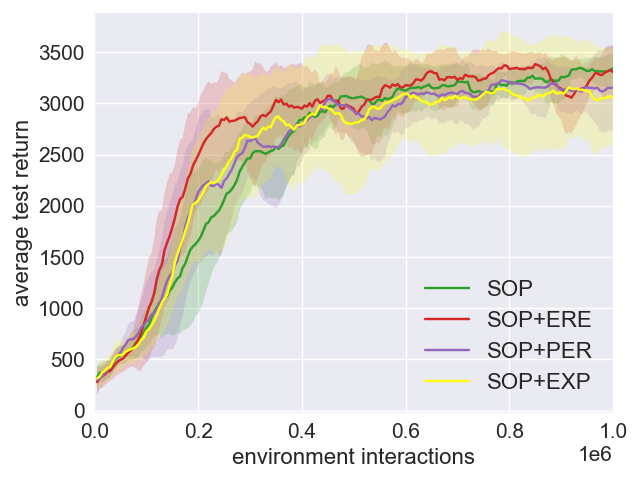}
	\caption{Hopper-v2}
\end{subfigure}
\begin{subfigure}{0.3\textwidth}
	\centering
	\includegraphics[width=0.95\linewidth]{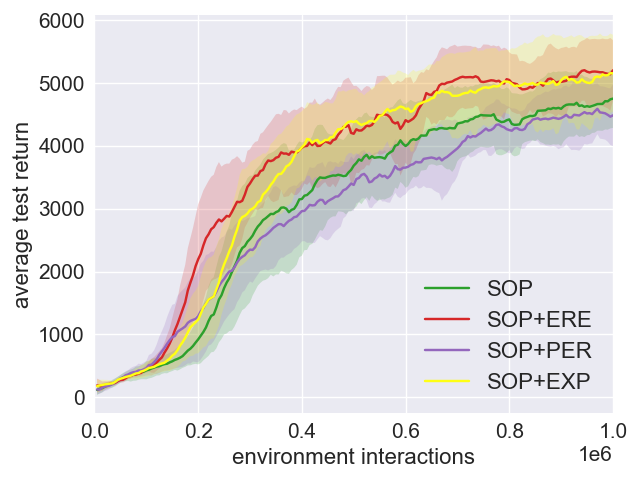}
	\caption{Walker2d-v2}
\end{subfigure}
\begin{subfigure}{0.3\textwidth}
	\centering
	\includegraphics[width=0.95\linewidth]{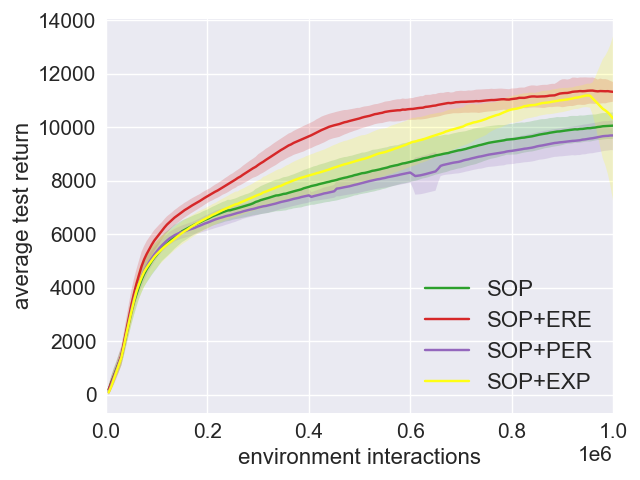}
	\caption{Halfcheetah-v2}
\end{subfigure}
\begin{subfigure}{0.3\textwidth}
	\centering
	\includegraphics[width=0.95\linewidth]{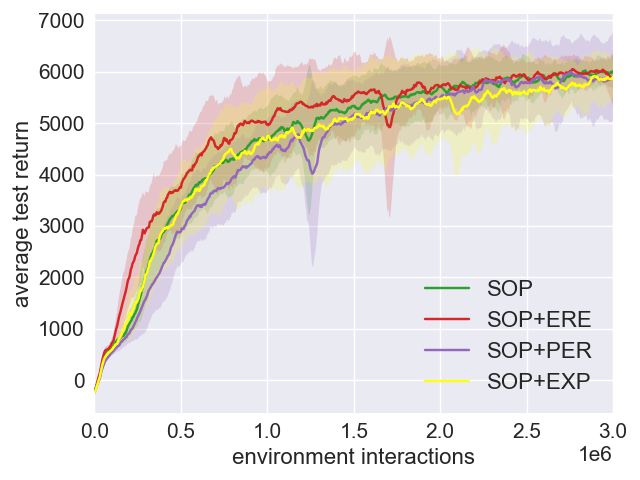}
	\caption{Ant-v2}
\end{subfigure}
\begin{subfigure}{0.3\textwidth}
	\centering
	\includegraphics[width=0.95\linewidth]{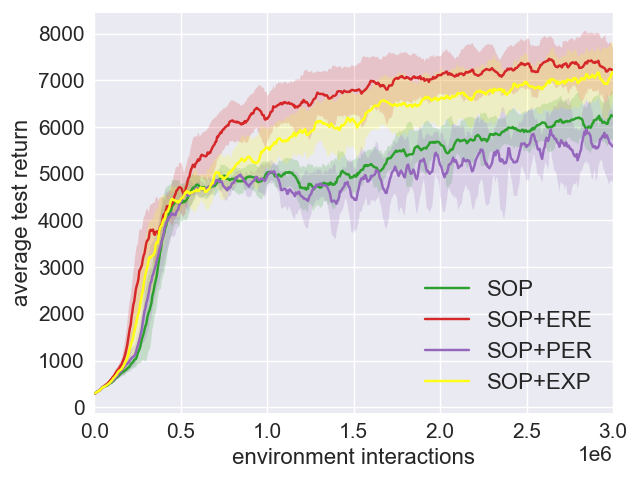}
	\caption{Humanoid-v2}
\end{subfigure}
\caption{Streamlined Off-Policy (SOP), with ERE and PER sampling schemes}
\label{fig:sop-per}
\end{figure*}

\subsection{SOP with Exponential Sampling}
The ERE scheme is similar to an exponential sampling scheme where we assign the probability of sampling according to the probability density function of an exponential distribution. Essentially, in such a sampling scheme, the more recent data points get exponentially more probability of being sampled compared to older data. 

For the $i^{th}$ most recent data point, the probability of sampling a data point $P(i)$ is computed as:
\begin{equation}
    P(i) = \lambda e^{-\lambda x}
\end{equation}

We apply this sampling scheme to SOP and refer to this variant as SOP+EXP. 

\subsection{PER and EXP experiment results}
Figure \ref{fig:sop-per} shows a performance comparison of SOP, SOP+ERE, SOP+EXP and SOP+PER. Results show that the exponential sampling scheme gives a boost to the performance of SOP, and especially in the Humanoid environment, although not as good as ERE. Surprisingly, SOP+PER does not give a significant performance boost to SOP (if any boost at all). We also found that it is difficult to find hyperparameter settings for SOP+PER that work well for all environments. Some of the other hyperparameter settings actually reduce performance. It is unclear why PER does not work so well for SOP. A similar result has been found in another recent paper \citep{fu2019diagnosing}, showing that PER can significantly reduce performance on TD3. Further research is needed to understand how PER can be successfully adapted to environments with continuous action spaces and dense reward structure. 

\section{Additional ERE analysis}

\begin{figure*}[!t]
\centering
        \begin{subfigure}[b]{0.325\textwidth}
                \includegraphics[width=\linewidth]{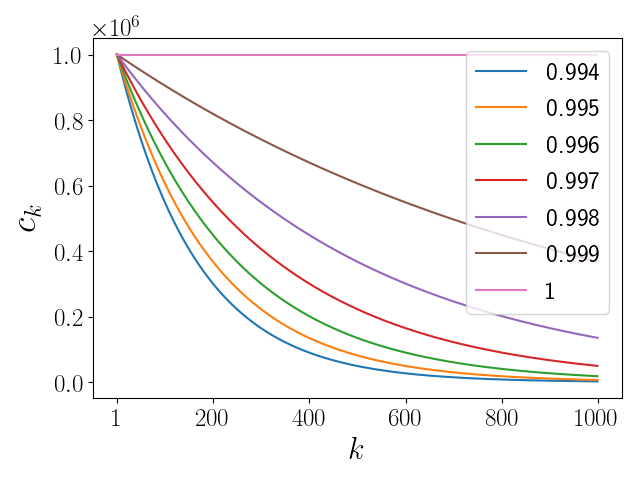}
                \caption{}
                \label{fig:eta_effect}
        \end{subfigure}%
        \begin{subfigure}[b]{0.325\textwidth}
                \includegraphics[width=\linewidth]{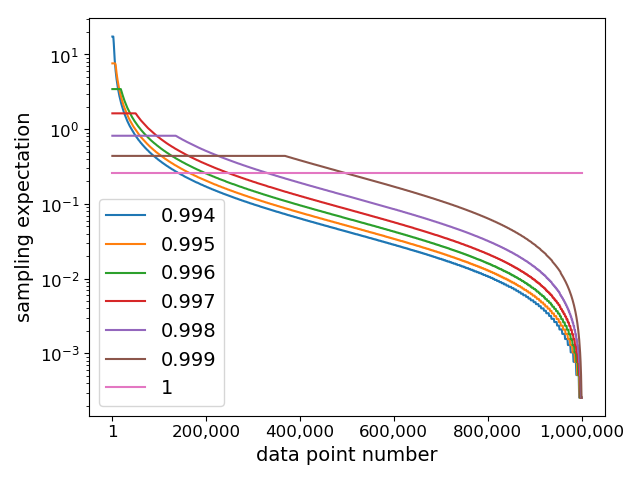}
                \caption{}
                \label{fig:data_expect}
        \end{subfigure}%
        \caption{Effect of different $\eta$ values. The plots assume a replay buffer with 1 million samples, and 1,000 mini-batches of size 256 in an update phase. Figure \ref{fig:eta_effect} plots $c_k$ (ranging from 0 to 1 million) as a function of $k$ (ranging from 1 to 1,000). Figure \ref{fig:data_expect} plots the expected number of times a data point in the buffer is sampled, with the data points ordered from most to least recent.}
        \label{fig:ere_effect}
\end{figure*}

Figure \ref{fig:ere_effect} shows, for fixed $\eta$, how $\eta$ affects the data sampling process, under the ERE sampling scheme. Recent data points have a much higher probability of being sampled compared to older data, and a smaller $\eta$ value gives more emphasis to recent data. 

Different $\eta$ values are desirable depending on how fast the agent is learning and how fast the past experiences become obsolete. So to make ERE work well in different environments with different reward scales and learning progress, we adapt $\eta$ to the the speed of learning. To this end, define performance to be the training episode return. Define $I_{recent}$ to be how much performance improved from $N/2$ timesteps ago, and $I_{max}$ to be the maximum improvement throughout training, where $N$ is the buffer size. Let the hyperparameter $\eta_0$ be the initial $\eta$ value. We then adapt $\eta$ according to the formula: 
$\eta = \eta_0 \cdot I_{recent}/I_{max} + 1 - (I_{recent}/I_{max})$. 

Under such an adaptive scheme, when the agent learns quickly, the $\eta$ value is low in order to learn quickly from new data. When progress is slow, $\eta$ is higher to make use of the stabilizing effect of uniform sampling from the whole buffer. 

\section{Additional implementation details}
\subsection{ERE implementation}
In this section we discuss some programming details. These details are not necessary for understanding the algorithm, but they might help with reproducibility.

In the ERE scheme, the sampling range always starts with the entire buffer (1M data) and then gradually shrinks. This is true even when the buffer is not full. So even if there are not many data points in the buffer, we compute $c_k$ based as if there are 1M data points in the buffer. One can also modify the design slightly to obtain a variant that uses the current amount of data points to compute $c_k$. In addition to the reported scheme, we also tried shrinking the sampling range linearly, but it gives less performance gain. 

In our implementation we set the number of updates after an episode to be the same as the number of timesteps in that episode. Since environments do not always end at 1000 timesteps, we can give a more general formula for $c_k$. Let $K$ be the number of mini-batch updates, let $N$ be the max size of the replay buffer, then:

\begin{equation}
c_k = N \cdot \eta^{k\frac{1000}{K}} 
\end{equation}

With this formulation, the range of sampling shrinks in more or less the same way with varying number of mini-batch updates. We always do uniform sampling in the first update, and we always have $\eta^{K\frac{1000}{K}} = \eta^{1000}$ in the last update.

When $\eta$ is small, $c_k$ can also become small for some of the mini-batches. To prevent getting a mini-batch with too many repeating data points, we set 
the minimum value for $c_k$ to 5000. We did not find this value to be too important and did not find the need to tune it. It also does not have any effect for any $\eta \geq 0.995$ since the sampling range cannot be lower than 6000. 

In the adaptive scheme with buffer of size 1M, the recent performance improvement is computed as the difference of the current episode return compared to the episode return 500,000 timesteps earlier. Before we reach 500,000 timesteps, we simply use $\eta_0$. The exact way of computing performance improvement does not have a significant effect on performance as long as it is reasonable. 

\begin{figure*}[!ht]
\centering
\begin{subfigure}{0.3\textwidth}
	\centering
	\includegraphics[width=0.95\linewidth]{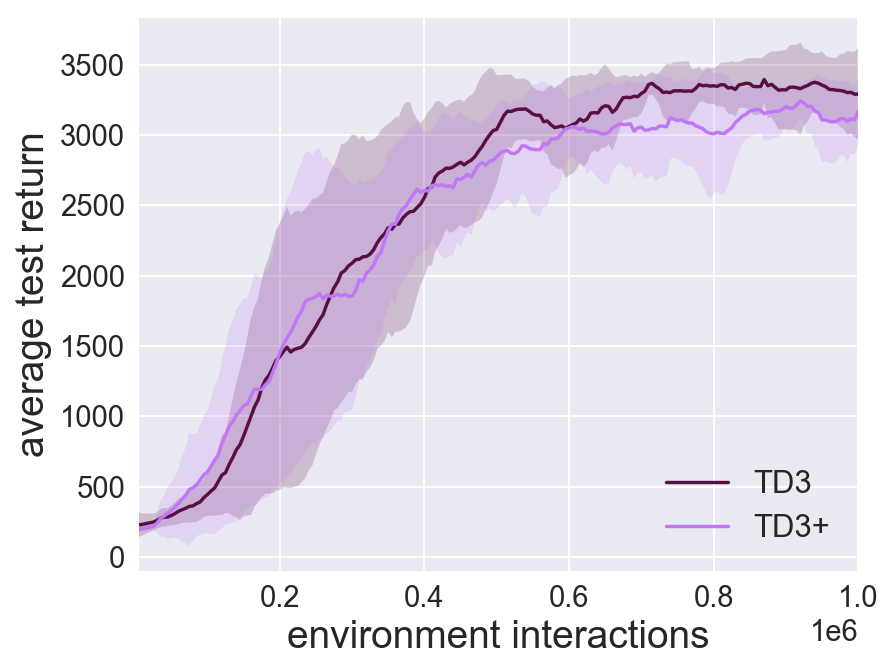}
	\caption{Hopper-v2}
	\label{fig:td3-hopper}
\end{subfigure}
\begin{subfigure}{0.3\textwidth}
	\centering
	\includegraphics[width=0.95\linewidth]{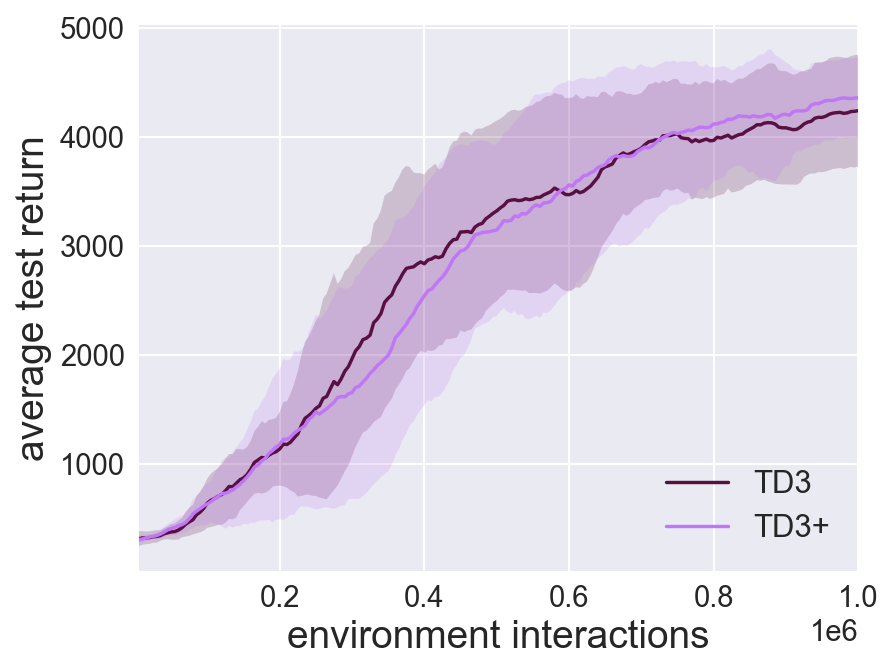}
	\caption{Walker2d-v2}
    \label{fig:td3-walker2d}
\end{subfigure}
\begin{subfigure}{0.3\textwidth}
	\centering
	\includegraphics[width=0.95\linewidth]{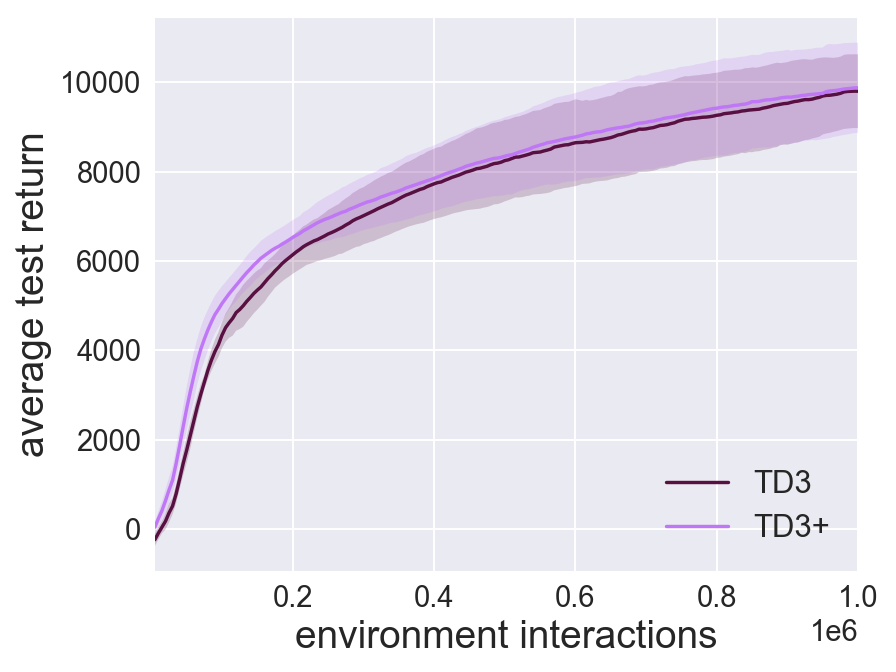}
	\caption{HalfCheetah-v2}
	\label{fig:td3-halfcheetah}
\end{subfigure}
\begin{subfigure}{0.3\textwidth}
	\centering
	\includegraphics[width=0.95\linewidth]{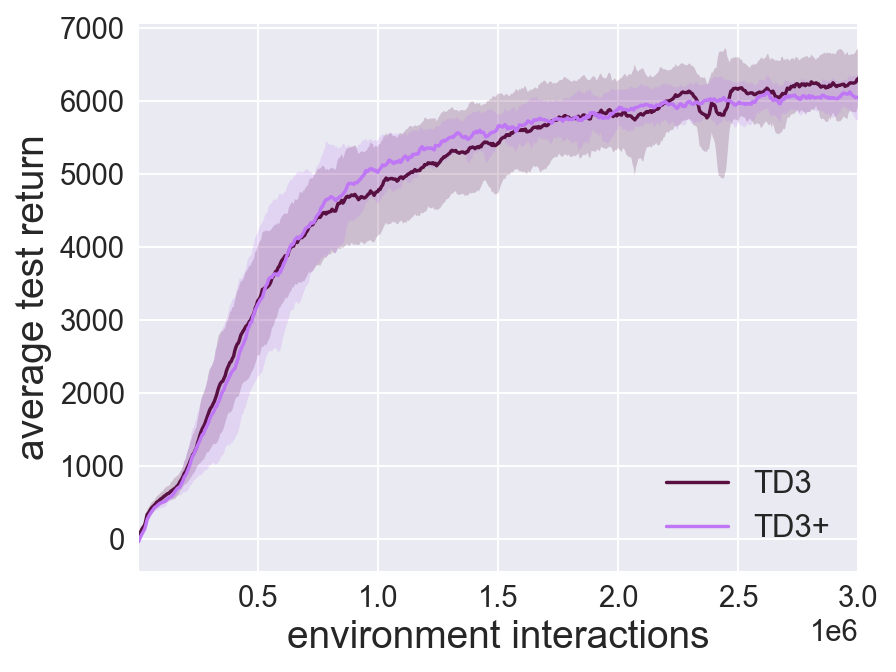}
	\caption{Ant-v2}
	\label{fig:td3-ant}
\end{subfigure}
\begin{subfigure}{0.3\textwidth}
	\centering
	\includegraphics[width=0.95\linewidth]{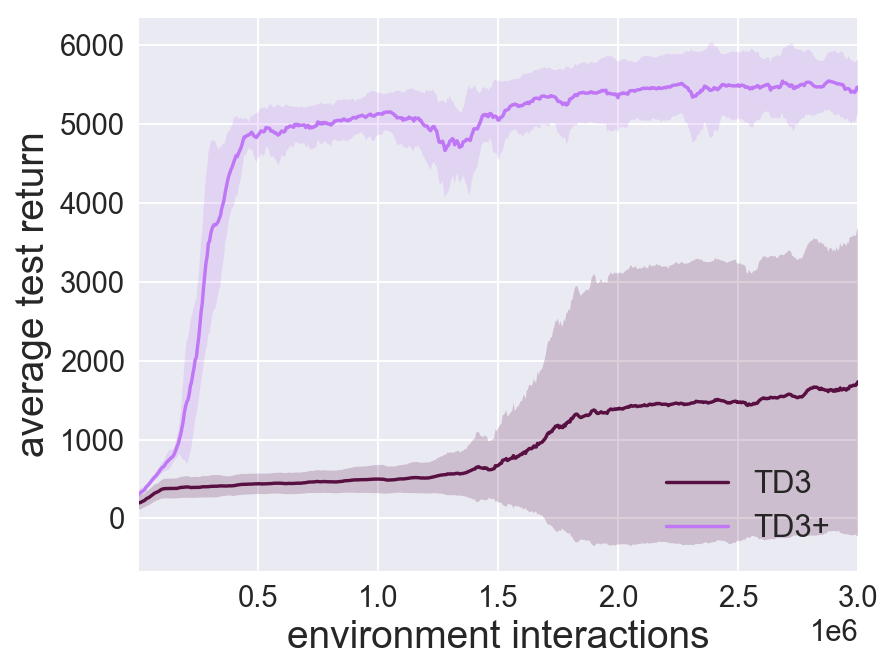}
	\caption{Humanoid-v2}
	\label{fig:td3-humanoid}
\end{subfigure}
\caption{TD3 versus TD3+ (TD3 plus the normalization scheme)}
\label{fig:TD3}
\end{figure*}

\subsection{Programming and computation complexity}
In this section we give analysis on the additional programming and computation complexity brought by ERE and PER. 

In terms of programming complexity, ERE is a clear winner since it only requires a small adjustment to how we sample mini-batches. It does not modify how the buffer stores the data, and does not require a special data structure to make it work efficiently. Thus the implementation difficulty is minimal. PER (proportional variant) requires a sum-tree data structure to make it run efficiently. The implementation is not too complicated, but compared to ERE it is a lot more work.

The exponential sampling scheme is very easy to implement, although a naive implementation will incur a significant computation overhead when sampling from a large buffer. To improve its computation efficiency, we instead uses an approximate sampling method. We first sample data indexes from segments of size 100 from the replay buffer, and then for each segment sampled, we sample one data point uniformly from that segment. 

In terms of computation complexity (not sample efficiency), and wall-clock time, ERE's extra computation is negligible. In practice we observe no difference in computation time between SOP and SOP+ERE. PER needs to update the priority of its data points constantly and compute sampling probabilities for all the data points. The complexity for sampling and updates is $O(log(N))$, and the rank-based variant is similar \citep{schaul2015prioritized}. Although this is not too bad, it does impose a significant overhead on SOP: SOP+PER runs twice as long as SOP. Also note that this overhead grows linearly with the size of the mini-batch. The overhead for the MuJoCo environments is higher compared to Atari, possibly because the MuJoCo environments have a smaller state space dimension while a larger batch size is used, making PER take up a larger portion of computation cost. For the exponential sampling scheme, the extra computation is also close to negligible when using the approximate sampling method. 

In terms of the proposed normalization scheme and the Inverting Gradients (IG) method, the normalization is very simple and can be easily implemented and has negligible computation overhead. IG has a simple idea, but its implementation is slightly more complicated than the normalization scheme. When implemented naively, IG can have a large computation overhead, but it can be largely avoided by making sure the gradient computation is still done in a batch-manner. We have made a very efficient implementation and our code is publicly available so that interested reader can easily reproduce it. 

\subsection{Computing Infrastructure}
Experiments are run on cpu nodes only. Each job runs on a single Intel(R) Xeon(R) CPU E5-2620 v3 with 2.40GHz. 

%\section{Additional experimental results}

\begin{comment}
\subsection{Inverting Gradients with ERE}
In Figure \ref{fig:ig-ere} we show additional results on applying ERE to SOP+IG. The result shows that after applying the ERE scheme, SOP and IG both get a performance boost. The performance of the SOP+ERE and IG+ERE are similar. 

\begin{figure*}[!htb]
\centering
\begin{subfigure}{0.3\textwidth}
	\centering
	\includegraphics[width=0.95\linewidth]{Figures/igere_hopper.png}
	\caption{Hopper-v2}
\end{subfigure}
\begin{subfigure}{0.3\textwidth}
	\centering
	\includegraphics[width=0.95\linewidth]{Figures/igere_walker2d.png}
	\caption{Walker2d-v2}
\end{subfigure}
\begin{subfigure}{0.3\textwidth}
	\centering
	\includegraphics[width=0.95\linewidth]{Figures/igere_halfcheetah.png}
	\caption{Halfcheetah-v2}
\end{subfigure}
\begin{subfigure}{0.3\textwidth}
	\centering
	\includegraphics[width=0.95\linewidth]{Figures/igere_ant.png}
	\caption{Ant-v2}
\end{subfigure}
\begin{subfigure}{0.3\textwidth}
	\centering
	\includegraphics[width=0.95\linewidth]{Figures/igere_humanoid.png}
	\caption{Humanoid-v2}
\end{subfigure}
\caption{SOP and inverting gradients with ERE sampling scheme}
\label{fig:ig-ere}
\end{figure*}
\end{comment}

\section{TD3 versus TD3+}
In figure \ref{fig:TD3}, we show additional results comparing TD3 with TD3 plus our normalization scheme, which we refer as TD3+.
The results show that after applying our normalization scheme, TD3+ has a significant performance boost in Humanoid, while in other environments, both algorithms achieve similar performance.

\end{document}